\pdfoutput=1
\documentclass[11pt]{article}
\usepackage{acl}

\usepackage{times}
\usepackage{latexsym}
\usepackage[T1]{fontenc}
\usepackage[utf8]{inputenc}
\usepackage{microtype}
\usepackage{booktabs}
\usepackage{graphicx}
\usepackage{xcolor}

\definecolor{darkblue}{rgb}{0, 0, 0.5}
\hypersetup{colorlinks=true, citecolor=darkblue, linkcolor=darkblue, urlcolor=darkblue}

\newcommand{\bench}{\textsc{MobileJudgeBench}}

\newcommand{\eg}{\textit{e.g.}}

\title{Benchmarking LLM Judges for Mobile Agent Evaluation}

\author{
 \textbf{Ziqiang Wang\textsuperscript{1,2}},
 \textbf{Li Gu\textsuperscript{1,2}},
 \textbf{Zhixiang Chi\textsuperscript{3}},
 \textbf{Zhi Liu\textsuperscript{4}},
\\
 \textbf{Seyed Mehdi Ayyoubzadeh\textsuperscript{5}},
 \textbf{Yuanhao Yu\textsuperscript{5}},
 \textbf{Yang Wang\textsuperscript{1,2}}
\\
\\
 \textsuperscript{1}Concordia University,
 \textsuperscript{2}Mila – Québec AI Institute,
 \textsuperscript{3}University of Toronto, \\
 \textsuperscript{4}Shanghai University,
 \textsuperscript{5}McMaster University
}

\begin{document}

\maketitle

\begin{abstract}
    Mobile agent benchmarks increasingly rely on LLM-based judges to evaluate task completion, yet the reliability of these judges on mobile agent trajectories remains largely unexamined. We introduce \bench{},\footnote{Data: \url{https://huggingface.co/datasets/WesleyWzq/MobileJudgeBench}. Code: \url{https://github.com/WZq975/MobileJudgeBench}.} a benchmark for systematically evaluating LLM-as-a-judge methods on mobile agent trajectories. Our benchmark comprises 931 human-annotated trajectories spanning 6 mobile agent benchmarks, 4 agent models, and 59 apps. We evaluate 6 judge methods (five adapted from SPA-Bench, A3 with two modes, AndroidArena, and AgentRewardBench, plus a simple baseline we design) across multiple LLM backbones. Our experiments reveal three key findings. First, a simple baseline judge with sampled screenshots is competitive with, and often exceeds, purpose-built methods, indicating that more elaborate judge pipelines do not consistently improve judge quality. Among competitive methods, the LLM backbone is the primary driver. Second, benchmark quality metrics are associated with judge utility in the two settings we study: they correlate with agent ranking fidelity for evaluation, and they track downstream performance when judges serve as reward signals for on-policy reinforcement learning. Third, failure analysis across two LLM backbones uncovers qualitatively opposite failure profiles, one conservative and the other permissive, linked to the backbone's precision-recall characteristics.
\end{abstract}
    
    \section{Introduction}
    \label{sec:intro}
    
    Using LLMs to judge the outputs of other models has emerged as a scalable alternative to human evaluation across a wide range of AI tasks~\citep{zheng2023judging, gu2024survey}. In the domain of autonomous mobile agents that operate smartphones on behalf of users, LLM-based judges play a dual role: they serve as \emph{evaluators} in benchmarks that measure agent progress on realistic Android tasks~\citep{rawles2025androidworld, chen2025spabench, chai2025a3, xing2024androidarena, lee2024bmoca, xu2024androidlab}, and increasingly as \emph{reward signals} for reinforcement learning (RL) training~\citep{bai2024digirl, qi2025webrl, mobilerl2025}. Judging mobile agent trajectories is particularly challenging: it requires interpreting multimodal evidence (sequences of screenshots, UI element trees, and executed actions) across long trajectories spanning diverse apps and interaction patterns. This makes judge quality critical not only for measuring progress but also for driving it.
    
    While some benchmarks provide rule-based state checkers~\citep{rawles2025androidworld, xu2024androidlab}, these require per-task engineering and are difficult to scale to new tasks or apps. As a result, benchmarks increasingly adopt LLM-based judges, whether through coarse-to-fine screenshot matching~\citep{chen2025spabench}, essential-state decomposition~\citep{chai2025a3}, or direct LLM judgment~\citep{xing2024androidarena}. However, the reliability of these LLM-based evaluators on mobile agent trajectories has not been systematically examined. Researchers select a benchmark, adopt its built-in judge, and report numbers without questioning the evaluation itself. This is problematic for two reasons. First, an unreliable judge produces noisy leaderboards: a judge that misclassifies even 10--15\% of trajectories can shift agent rankings substantially (\S\ref{sec:meta_corr}). Second, when judges serve as reward signals for RL training, errors in the reward directly corrupt the learning process~\citep{gao2023scaling, huang2024darkside}. Yet neither the degree of unreliability nor which quality metrics (\eg, accuracy, precision, recall) matter most for these downstream applications has been established for mobile agents. Existing judge benchmarks target web agents~\citep{lu2025agentrewardbench}, whereas mobile judges rely on Android UI trees and action-marked screenshots, and a judge fine-tuned on web trajectories is less accurate on our data than every judge we evaluate (\S\ref{sec:judge_eval}).
    
\begin{figure*}[t]
    \centering
    \includegraphics[width=0.91\textwidth]{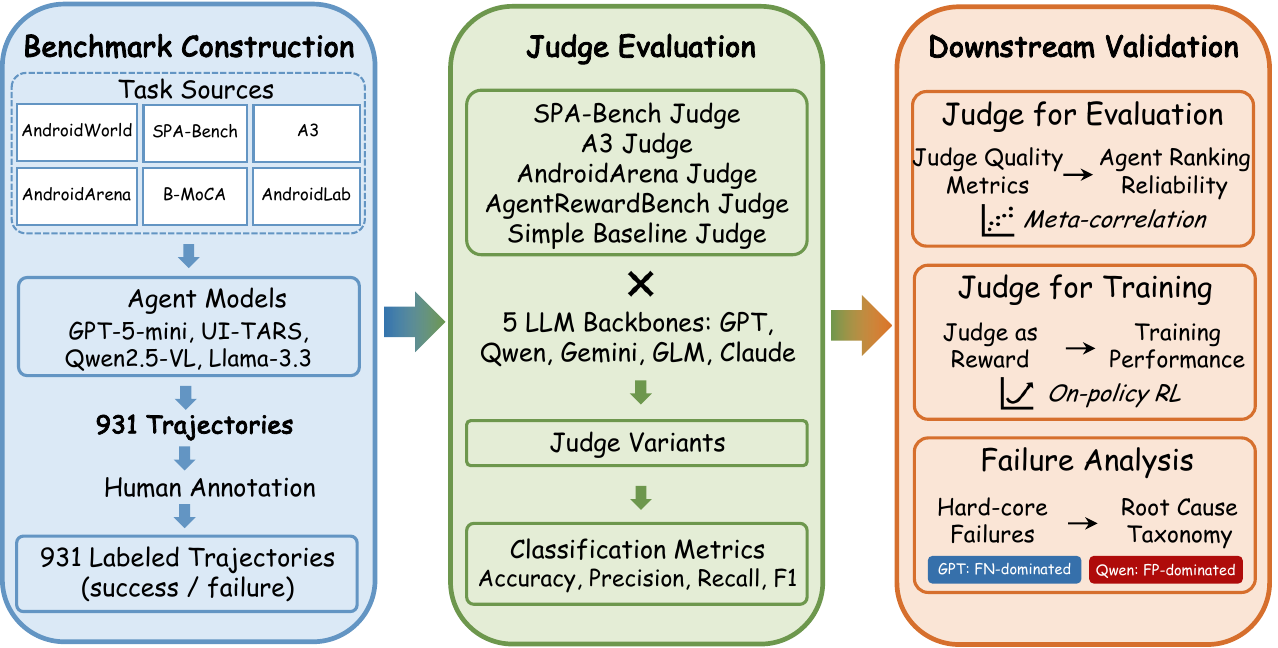}
    \caption{Overview of \bench{}. We construct a judge benchmark from 931 human-annotated mobile agent trajectories spanning 6 benchmarks (left, \S\ref{sec:benchmark}), evaluate 6 LLM-as-a-judge methods across 5 backbones (middle, \S\ref{sec:judge_eval}), and relate benchmark metrics to judge utility for agent evaluation (\S\ref{sec:meta_corr}) and on-policy training (\S\ref{sec:training}), complemented by a root-cause failure analysis (\S\ref{sec:failure}) (right).}
    \label{fig:overview}
    \end{figure*}

    In this work, we introduce \textsc{Mobile\-Judge\-Bench}, to our knowledge the first benchmark for evaluating LLM-as-a-judge methods on mobile agent trajectories (Figure~\ref{fig:overview}). We collect 931 trajectories from 6 established mobile agent benchmarks, generated by 4 diverse agent models across 59 apps, and obtain human expert annotations with multi-annotator redundancy. Using this benchmark, we evaluate 6 judge methods (five adapted from SPA-Bench, A3 with two modes, AndroidArena, and AgentRewardBench, plus a simple baseline we design for controlled ablation) across multiple LLM backbones. Crucially, we go beyond measuring intrinsic judge accuracy: we test whether our benchmark metrics are associated with judge utility in two downstream uses, agent evaluation and RL training.
    
    Our contributions are as follows:
    \begin{enumerate}
        \item \textbf{A judge benchmark for mobile agents.} We construct a dataset of 931 human-annotated trajectories spanning 6 benchmarks, 4 agents, and 59 apps, together with a unified evaluation framework that standardizes judge assessment with classification and ranking metrics (\S\ref{sec:benchmark}).
    
        \item \textbf{Systematic evaluation of judge methods.} We evaluate 6 judge methods across 5 LLM backbones, revealing that no single method dominates and that judge accuracy depends substantially on both the method and the LLM backbone (\S\ref{sec:judge_eval}).
    
        \item \textbf{A simple baseline judge with ablation study.} We design a streamlined judge that is competitive with or exceeds purpose-built methods (up to 90.9\% accuracy). Ablations over screenshot count, image resolution, UI metadata, and agent reasoning show that screenshot count is the dominant design variable, while the other inputs have marginal impact (\S\ref{sec:simple_baseline}, \S\ref{sec:ablation}).
    
        \item \textbf{Linking benchmark metrics to downstream judge utility.} We examine two downstream applications. For \emph{evaluation}, meta-correlation analysis shows that accuracy, F1, and balanced accuracy are strongly associated with agent ranking fidelity and success-rate estimation accuracy, whereas precision is not associated with ranking fidelity. For \emph{training}, on-policy RL experiments in AndroidWorld show that the ordering of judges by benchmark accuracy matches the ordering of the agents trained with them (\S\ref{sec:applications}).
    
        \item \textbf{Failure pattern analysis.} We analyze hard-core failure cases where nearly all judge methods err, revealing that different LLM backbones produce qualitatively opposite failure profiles, one conservative (false-negative-heavy) and the other permissive (false-positive-heavy), with distinct root cause taxonomies (\S\ref{sec:failure}).
    \end{enumerate}
    
    \section{Related work}
    \label{sec:related}
    
    \noindent{\bf LLM-as-a-judge.}
    Using LLMs as surrogate evaluators was popularized by MT-Bench and Chatbot Arena~\citep{zheng2023judging}, where GPT-4 achieves over 80\% agreement with human preferences on chat evaluation. Subsequent work has identified systematic biases in LLM judges, including position bias, verbosity bias, and self-preference~\citep{gu2024survey, tan2024llms}, and studied judge reliability~\citep{dong2025trust}, contamination via preference leakage~\citep{li2025preference}, and fine-tuning dedicated judge models~\citep{zhu2024judgelm, mcaleese2024criticgpt}. \citet{zhuge2024agent} extend the paradigm to Agent-as-a-Judge, equipping evaluators with agentic capabilities. This body of work focuses predominantly on evaluating text outputs (\eg, chat responses, code). In the mobile agent domain, LLM judges must handle multi-step trajectories with interleaved screenshots, actions, and UI states, as seen in methods from SPA-Bench~\citep{chen2025spabench}, A3~\citep{chai2025a3}, and AndroidArena~\citep{xing2024androidarena}. We test for self-preference in this setting in Appendix~\ref{app:self_judging}.

    \noindent{\bf Mobile and GUI agent benchmarks.}
    The mobile agent evaluation landscape is fragmented across benchmarks that differ in task design, environment, and evaluation method. AndroidWorld~\citep{rawles2025androidworld} provides programmatic, state-based success checking; B-MoCA~\citep{lee2024bmoca} tests generalization across device configurations with rule-based detectors; SPA-Bench~\citep{chen2025spabench} uses coarse-to-fine screenshot matching with MLLM verification; A3~\citep{chai2025a3} employs essential-state decomposition; AndroidArena~\citep{xing2024androidarena} uses direct GPT-4 judgment; and AndroidLab~\citep{xu2024androidlab} verifies sub-goal completion via UI tree matching. On the desktop and web side, OSWorld~\citep{xie2024osworld}, WebArena~\citep{zhou2024webarena}, VisualWebArena~\citep{koh2024visualwebarena}, and Mind2Web~\citep{deng2023mind2web} provide complementary evaluation environments. Large-scale datasets such as Android-in-the-Wild~\citep{rawles2023aitw} and AndroidControl~\citep{li2024androidcontrol} supply training data. While rule-based checkers are reliable by construction for their supported tasks, the LLM-based judges that many benchmarks rely on, from SPA-Bench to AndroidDaily~\citep{sui2026androiddaily}, have not been systematically validated against human ground truth.

    \noindent{\bf Automatic evaluation of agent trajectories.}
    AgentRewardBench~\citep{lu2025agentrewardbench} is the most directly related work: it benchmarks LLM judges on 1,302 web agent trajectories across 5 web benchmarks, evaluating judge methods including AER~\citep{pan2024aer} and NNetNav~\citep{murty2024nnetnav}. Concurrent work on mobile evaluation includes AutoEval~\citep{autoeval2025}, which decomposes tasks into expected success states for AndroidLab. WebJudge~\citep{xu2025webjudge} proposes a three-stage judge pipeline for web agent trajectories and releases a fine-tuned 7B judge model. We evaluate both on our benchmark (Appendix~\ref{app:webjudge}). More recent approaches move beyond passive screenshot inspection: VAGEN~\citep{vagen2026} and ProRe~\citep{prore2025} introduce proactive verification, where the judge actively interacts with the environment to collect evidence, and AJ-Bench~\citep{shi2026ajbench} evaluates such environment-aware judges on search, data-system, and desktop-GUI tasks. Our work differs from AgentRewardBench in three key respects: we focus on mobile agents with their distinct observation types and fragmented benchmark ecosystem; we integrate judge methods from 4 existing agent benchmarks into a unified evaluation; and we empirically link judge quality to downstream on-policy training performance, a connection not explored in existing agent judge benchmarks (Appendix~\ref{app:arb_comparison}).

    \noindent{\bf Reward modeling for agent training.}
    The use of learned or model-based rewards for training originates from RLHF~\citep{ouyang2022instructgpt} and has been extended through DPO~\citep{rafailov2023dpo} and RLAIF~\citep{bai2022constitutional}. In the agent domain, DigiRL~\citep{bai2024digirl} demonstrates that a VLM evaluator can serve as the reward signal for RL training of device-control agents, achieving substantial improvements over supervised fine-tuning. WebRL~\citep{qi2025webrl} shows that reward model quality is critical for web agent training. For mobile agents specifically, MobileRL~\citep{mobilerl2025} and MobileGUI-RL~\citep{mobileguirl2025} apply online RL with various reward signals. RewardBench~\citep{lambert2025rewardbench} provides a general benchmark for reward models but does not cover agent-specific evaluation. Scaling laws for reward model overoptimization~\citep{gao2023scaling} establish that imperfect reward models lead to predictable performance degradation. Related to this line of work, we empirically measure whether judge quality metrics on a benchmark are reflected in downstream agent training performance.

    \section{Judge benchmark construction}
    \label{sec:benchmark}
    
    \bench{} consists of 931 human-annotated mobile agent trajectories drawn from 6 established benchmarks, generated by 4 agent models across 289 unique tasks and 59 real Android apps. We describe the data collection, annotation process, and evaluation framework below.
    
    \subsection{Task and trajectory collection}
    \label{sec:data}
    
    \noindent{\bf Task sources.}
    We select tasks from 6 mobile agent benchmarks (Table~\ref{tab:benchmark_stats}) that collectively represent the major evaluation paradigms: programmatic state checking (AndroidWorld, B-MoCA), screenshot matching (SPA-Bench), essential-state decomposition (A3), direct LLM judgment (AndroidArena), and sub-goal matching (AndroidLab).

    \vspace{-1mm}
    \noindent{\bf Agent models and trajectory format.}
    We generate trajectories using 4 agents: GPT-5-mini and Qwen2.5-VL-72B (via M3A~\citep{rawles2025androidworld}), UI-TARS-72B~\citep{uitars2025}, and Llama-3.3-70B~\citep{dubey2024llama3} (via T3A). Each agent attempts every task, yielding up to 4 trajectories per task. Each trajectory is a sequence of (screenshot, action, UI tree, agent reasoning) tuples recorded at every step, stored in a unified format.
    
    \begin{table}[t]
    \centering
    \small
    \begin{tabular}{lrrr}
    \toprule
    \textbf{Benchmark} & \textbf{Tasks} & \textbf{Traj.} & \textbf{Apps} \\
    \midrule
    SPA-Bench     & 69 & 220 & 33 \\
    AndroidWorld  & 52 & 205 &  9 \\
    A3            & 58 & 154 & 19 \\
    AndroidArena  & 51 & 151 & 13 \\
    B-MoCA        & 35 & 123 & 12 \\
    AndroidLab    & 24 &  78 &  7 \\
    \midrule
    \textbf{Total} & \textbf{289} & \textbf{931} & \textbf{59} \\
    \bottomrule
    \end{tabular}
    \caption{Benchmark statistics. Tasks are unique task definitions, and trajectories include up to 4 agent runs per task. Apps are distinct applications, including every app involved in a cross-app task. The total is smaller than the column sum because some apps appear in multiple benchmarks.}
    \label{tab:benchmark_stats}
    \end{table}
    
    \subsection{Human annotation}
    \label{sec:annotation}
    
    We recruit 9 graduate-student annotators to label each trajectory with a binary success judgment (success/failure). Each trajectory is independently annotated by 2--4 annotators using a custom annotation platform (Appendix~\ref{app:benchmark}), which provides trajectory video playback, step-by-step screenshot inspection, and structured annotation forms. Annotators review the task instruction, screenshots, actions, and agent reasoning. Average pairwise agreement on the success label is 88.4\%. For trajectories with annotator disagreement, annotators discuss and finalize the label. The final dataset is approximately balanced: 492 success (52.8\%) and 439 failure (47.2\%).
    
    \subsection{Evaluation framework}
    \label{sec:eval_framework}
    
    We design a unified evaluation pipeline that takes any judge method and any trajectory as input and produces standardized metrics. Given a set of judge predictions and human ground-truth labels, we report accuracy, precision, recall, F1, and balanced accuracy at the trajectory level. A ``positive'' is a successful trajectory: precision measures how often the judge's success predictions are correct, while recall measures how often truly successful trajectories are identified. Because the dataset is approximately balanced (53\% positive), accuracy and balanced accuracy are close, but we report both for completeness.

    \section{Judge methods}
    \label{sec:methods}
    
    We evaluate 6 LLM-as-a-judge methods on our benchmark. Five are adapted from existing agent benchmarks (SPA-Bench, A3 with two evaluation modes, AndroidArena, and AgentRewardBench) and one is a simple baseline we design for controlled ablation. All methods take a task instruction and agent trajectory as input and output a binary success/failure prediction with reasoning. They differ in how they represent the trajectory (screenshots vs.\ text, full sequence vs.\ final state) and how they structure the evaluation prompt.
    
    \subsection{Existing judge methods}
    \label{sec:existing_methods}
    
    \noindent{\bf SPA-Bench judge~\citep{chen2025spabench}.}
    The original SPA-Bench evaluation uses a coarse-to-fine pipeline: a coarse stage matches pre-annotated key components against the final screenshot, and a fine stage invokes an MLLM when the match is ambiguous. We use only the LLM-based fine stage, as the coarse stage requires task-specific key component annotations that are only available for SPA-Bench's own tasks. The judge receives all screenshots with annotated action markers (red dots at tap locations, scroll indicators) overlaid on the images, together with detailed evaluation guidelines.

    \noindent{\bf A3 judge~\citep{chai2025a3}.}
    A3 provides two evaluation modes. \emph{Final-state evaluation} presents only the last screenshot and its UI XML tree to the LLM, asking whether the task goal is achieved. \emph{Essential-states evaluation} first decomposes the task instruction into a set of necessary sub-goals (essential states) via an LLM call, then evaluates whether each essential state is achieved by examining screenshots within a sliding window. We evaluate both modes, with essential states generated by the LLM.

    \noindent{\bf AndroidArena judge~\citep{xing2024androidarena}.}
    This method formats the trajectory as a text-based chronological sequence of actions and structured UI element observations (extracted from accessibility trees). The formatted trajectory, together with the task instruction, is passed to an LLM for a direct success/failure judgment. The original method operates purely on text, and we add a screenshot-based fallback for trajectories that lack text observations. This is the only method that operates primarily on text.

    \noindent{\bf AgentRewardBench judge~\citep{lu2025agentrewardbench}.}
    Originally designed for web agent evaluation, this method evaluates trajectories along 4 dimensions via structured questions: (1)~task success, (2)~side effects, (3)~action optimality, and (4)~action looping. We adapt it for mobile agents by providing the last screenshot, the UI element list, and a step-by-step trajectory summary. The judge produces a reasoning trace followed by structured answers for each question, and we use the task success answer as the final prediction.

    \subsection{Simple baseline judge}
    \label{sec:simple_baseline}
    
    Existing methods entangle multiple design choices, making it difficult to isolate the effect of any single factor. We design a simple baseline judge with independently configurable components to enable controlled ablation. The judge receives the task instruction followed by a chronological step-by-step trajectory: each step includes the action taken, optionally the agent's reasoning and visible UI elements, and a screenshot uniformly sampled from the full trajectory. The system prompt provides balanced evaluation guidelines: outcome-focused judgment, mid-trajectory success recognition, corrective action credit, and balanced framing that avoids overly strict conditions which tend to suppress recall. By varying screenshot count (3--192), UI metadata, agent reasoning, and image resolution independently, we isolate the effect of each factor (\S\ref{sec:ablation}). The default configuration is 48 uniformly sampled screenshots at max long edge 600px, without UI metadata or agent reasoning.

    \section{Experiments}
    \label{sec:experiments}
    
    \subsection{Judge evaluation}
    \label{sec:judge_eval}
    
    We evaluate all 6 judge methods (5 existing + our simple baseline) across 5 LLM backbones: Qwen2.5-VL-72B~\citep{qwen2025qwen25vl}, GPT-5-mini~\citep{openai2025gpt5}, Gemini~3~Flash~\citep{googledeepmind2025gemini3flash}, GLM-4.6V~\citep{glmv2025}, and Claude Sonnet 4.5~\citep{anthropic2025claudesonnet45}. Table~\ref{tab:main_results} reports accuracy on the full 931-trajectory benchmark.
    
    \begin{table}[t]
    \centering
    \small
    \setlength{\tabcolsep}{2pt}  %
    \begin{tabular}{lccccc}
    \toprule
    \textbf{Judge Method} & \textbf{Qwen} & \textbf{GPT} & \textbf{Gemini} & \textbf{GLM} & \textbf{Claude} \\
    \midrule
    AndroidArena       & 84.6 & \textbf{88.2} & 87.7 & 83.9 & 79.0 \\
    A3 (final state)   & 81.5 & 76.4 & \textbf{83.8} & 78.1 & 78.4 \\
    A3 (essential)     & 80.0 & 78.1 & \textbf{83.7} & 81.9 & 81.3 \\
    SPA-Bench          & 80.1 & \textbf{90.6} & 89.8 & 82.2 & 84.7 \\
    AgentRewardBench   & 84.9 & 82.8 & \textbf{89.3} & \underline{86.1} & \underline{87.2} \\
    \midrule
    Simple Baseline$^\dagger$ & \underline{85.7} & \underline{90.8} & \underline{\textbf{90.9}} & 84.9 & 86.0 \\
    \bottomrule
    \end{tabular}
    \caption{Judge accuracy (\%) across methods and LLM backbones (Qwen2.5-VL-72B, GPT-5-mini, Gemini~3~Flash, GLM-4.6V, Claude Sonnet 4.5). Bold marks the best backbone per method (row-wise) and underline marks the best method per backbone (column-wise). Full results are in Appendix~\ref{app:full_results} and confidence intervals in Appendix~\ref{app:paired}. $^\dagger$Default configuration: 48 uniformly sampled screenshots at max long edge 600px, without UI metadata or agent reasoning (see ablation below).}
    \label{tab:main_results}
    \end{table}
    
    Several findings emerge from the results. First, \textbf{no single method dominates}: the simple baseline has the highest accuracy with the Gemini, GPT-5-mini, and Qwen backbones (90.9\%, 90.8\%, 85.7\%), while AgentRewardBench has the highest with GLM and Claude (86.1\% and 87.2\%). Second, \textbf{the LLM backbone substantially affects every method}: the same method can vary by 5.6--10.5pp across backbones (\eg, SPA-Bench ranges from 80.1\% with Qwen to 90.6\% with GPT). Third, \textbf{the simple baseline is competitive}: despite its minimal design, it is within 1.2pp of the strongest method even on the two backbones it does not lead, so more elaborate pipelines do not consistently improve quality. A two-way variance decomposition over the 6$\times$5 grid quantifies the two factors: method choice explains 49\% of the accuracy variance and the backbone 21\%, but the method share is driven by the two weakest methods, both purpose-built (the A3 modes). Excluding them as a sensitivity analysis reverses the shares (11\% vs.\ 49\%): among AndroidArena, SPA-Bench, AgentRewardBench, and the baseline, backbone choice dominates (Appendix~\ref{app:variance}).

    \vspace{-1mm}
    \noindent{\bf Uncertainty and cost.}
    We quantify the uncertainty of these comparisons with paired task-cluster bootstrap CIs, resampling tasks within each source benchmark (Appendix~\ref{app:paired}). The leader and the runner-up are statistically tied on every backbone: the baseline's margin over SPA-Bench is $+0.1$pp (95\% CI $[-1.8, +1.9]$) with GPT-5-mini and $+1.1$pp ($[-0.8, +2.8]$) with Gemini. Across its 25 comparisons with the purpose-built methods, the baseline is significantly more accurate in 15 and significantly less accurate in none. It is also inexpensive: with GPT-5-mini it uses 3.5k tokens per trajectory, compared with 21.9k for SPA-Bench and 18.8k for A3 essential-states (Appendix~\ref{app:cost}). Within each source benchmark, the baseline is within 3.8pp of the best method for every backbone (Appendix~\ref{app:per_benchmark}). We additionally evaluate WebJudge~\citep{xu2025webjudge}, a judge pipeline from the web domain, which scores 3.1pp below the baseline at 20$\times$ its token budget (Appendix~\ref{app:webjudge}).

    \vspace{-1mm}
    \noindent{\bf Ablation study.}
    \label{sec:ablation}
    We ablate the simple baseline along two dimensions (Table~\ref{tab:ablation}). For \emph{screenshot count vs.\ resolution}, at an equal pixel budget, fewer higher-resolution screenshots outperform many low-resolution ones: 12 screenshots at 1/4 resolution and 48 at 1/16 are within 1pp of each other on both backbones, while 192 at 1/64 is 3--4pp lower (86.7\%/82.5\%). For \emph{input components}, UI tree metadata has negligible impact (${\leq}$0.3pp for Qwen), and agent reasoning provides only a modest gain for GPT (${\sim}$1.6pp). With enough screenshots, additional metadata is unnecessary. We also run the baseline with only the final screenshot (about 1k tokens). It reaches 83.8/85.5/82.7/80.5\% with GPT/Gemini/GLM/Claude, compared with 76.4/83.8/78.1/78.4\% for A3 final-state, which sees the same frame plus its UI tree. Going from one to 48 frames adds 2.4--7.0pp (Appendix~\ref{app:final_frame}).
    
    \begin{table}[t]
    \centering
    \small
    \begin{tabular}{lcc}
    \toprule
    \textbf{Configuration} & \textbf{Qwen} & \textbf{GPT} \\
    \midrule
    \multicolumn{3}{l}{\textit{Screenshots $\times$ resolution (iso-budget, screenshots only)}} \\
    \quad 12 screenshots @ 1/4 res  & 86.7 & 90.3 \\
    \quad 48 screenshots @ 1/16 res & 85.7 & \textbf{90.8} \\
    \quad 192 screenshots @ 1/64 res& 82.5 & 86.7 \\
    \midrule
    \multicolumn{3}{l}{\textit{Input components (3 screenshots, original resolution)}} \\
    \quad + UI trees + reasoning    & 84.3 & 90.1 \\
    \quad + UI trees only           & 84.3 & 86.9 \\
    \quad + reasoning only          & 84.0 & 88.9 \\
    \quad screenshots only          & 84.0 & 87.3 \\
    \midrule
    \multicolumn{3}{l}{\textit{Final frame only (600px)}} \\
    \quad 1 screenshot              & --   & 83.8 \\
    \bottomrule
    \end{tabular}
    \caption{Ablation study on simple baseline judge. Top: fewer higher-resolution screenshots outperform many low-resolution ones at an equal pixel budget. Middle: UI metadata has minimal impact, and agent reasoning provides a modest gain for GPT only. Bottom: the baseline restricted to the final frame (Appendix~\ref{app:final_frame}).}
    \label{tab:ablation}
    \end{table}
    
    \subsection{Benchmark metrics and downstream judge utility}
    \label{sec:applications}
    
    A benchmark is only useful if its metrics relate to how judges perform in practice. We examine two downstream applications: using judges to \emph{evaluate} agents (ranking and success-rate estimation) and using judges as \emph{reward signals} for on-policy RL training.
    
    \subsubsection{Judge for evaluation: meta-correlation}
    \label{sec:meta_corr}
    
    For each of the 30 judge variants, we compute per-agent success rates (24 benchmark--model combinations) and compare against human ground truth via \emph{ranking fidelity} (Spearman~$\rho$) and \emph{rate estimation error} (MAE). We then compute a \emph{meta-correlation}: does higher judge quality predict more reliable agent evaluation? Details are in Appendix~\ref{app:full_results}.
    
\begin{figure*}[t]
    \centering
    \includegraphics[width=0.9\textwidth]{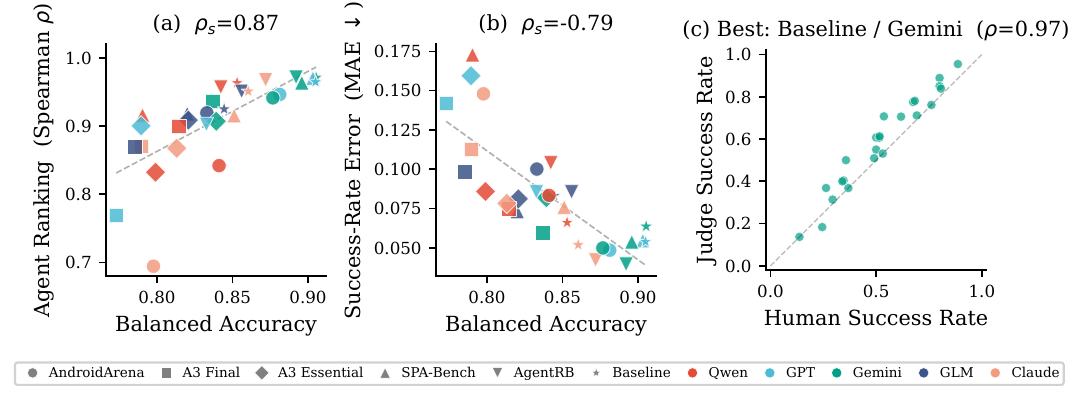}
    \vspace{-3mm}
    \caption{Judge quality and evaluation reliability (30 method$\times$backbone variants; marker shape$=$judge method, color$=$LLM backbone). (a)~Balanced accuracy correlates with agent ranking fidelity ($\rho_s{=}0.87$). (b)~Balanced accuracy inversely correlates with success-rate estimation error. (c)~The highest-accuracy judge (Baseline/Gemini, 90.9\%) closely tracks human success rates across 24 agents ($\rho{=}0.97$).}
    \label{fig:quality_vs_reliability}
    \end{figure*}

    Table~\ref{tab:meta_corr} and Figure~\ref{fig:quality_vs_reliability} present the results. F1 is the strongest predictor of ranking fidelity ($\rho_s = 0.90$, 95\% CI $[0.66, 0.92]$), while balanced accuracy best predicts rate estimation ($\rho_s = -0.79$ $[-0.92, -0.70]$). The CIs for accuracy, F1, and balanced accuracy exclude zero for both reliability metrics. Recall predicts ranking fidelity but is only weakly associated with rate error. Strikingly, precision has \emph{no predictive power} for ranking fidelity ($\rho_s = 0.00$) and at most a weak association with rate error. Evaluation reliability thus tracks balanced classification. The highest-accuracy judge (Baseline/Gemini, 90.9\%) closely tracks human success rates ($\rho = 0.97$, Figure~\ref{fig:quality_vs_reliability}c). The intervals are computed with a task-cluster bootstrap and corroborated by leave-one-out checks over backbones and methods (Appendix~\ref{app:variance}). We also test for self-preference, since two judge backbones share a model family with an agent. A difference-in-differences test finds no systematic leniency: pooled over methods, the Qwen judge is harsher on its own family's failed trajectories ($-8.8$pp false-positive rate, 95\% CI $[-11.6, -5.8]$) and the GPT-5-mini judge shows no difference ($+2.1$pp $[-3.4, +7.8]$; Appendix~\ref{app:self_judging}).
    
    \begin{table}[t]
    \centering
    \small
    \setlength{\tabcolsep}{2.5pt}
    \begin{tabular}{lcc}
    \toprule
    \textbf{Quality Metric} & \textbf{$\rho_s$ w/ Ranking} & \textbf{$\rho_s$ w/ MAE$\downarrow$} \\
    \midrule
    Accuracy      &  0.89 {\scriptsize$[0.66, 0.92]$}  & $-$0.77 {\scriptsize$[-0.92, -0.68]$} \\
    Precision     &  0.00 {\scriptsize$[-0.23, 0.30]$} & $-$0.22 {\scriptsize$[-0.36, 0.05]$} \\
    Recall        &  0.74 {\scriptsize$[0.43, 0.85]$}  & $-$0.33 {\scriptsize$[-0.64, -0.16]$} \\
    F1            &  \textbf{0.90} {\scriptsize$[0.66, 0.92]$}  & $-$0.72 {\scriptsize$[-0.89, -0.60]$} \\
    Balanced Acc  &  0.87 {\scriptsize$[0.65, 0.91]$}  & $-$\textbf{0.79} {\scriptsize$[-0.92, -0.70]$} \\
    \bottomrule
    \end{tabular}
    \caption{Meta-correlation across 30 judge variants. Each cell is the Spearman $\rho_s$ between a judge quality metric and a reliability metric: agent ranking fidelity ($\rho_s$ w/ Ranking) or success-rate estimation error ($\rho_s$ w/ MAE, where negative means higher quality $\to$ lower error). Brackets give 95\% CIs from a task-cluster bootstrap (tasks resampled with replacement within each source benchmark; 2{,}000 replicates; the 30 judge variants held fixed). Bold marks the strongest predictor per column.}
    \label{tab:meta_corr}
    \end{table}
    
    \subsubsection{Judge for training: on-policy RL}
    \label{sec:training}
    
    We train UI-TARS-7B-SFT~\citep{uitars2025} on the easy task set of AndroidWorld using GRPO with 4 reward configurations: the built-in rule-based checker, and our simple baseline judge with GPT-5-mini, GPT-5.2, and Qwen backbones. All conditions share identical hyperparameters and differ only in the reward source. Evaluation uses the ground-truth checker across 3 seeds (Appendix~\ref{app:training}).
    
\begin{figure*}[t]
    \centering
    \includegraphics[width=0.87\textwidth]{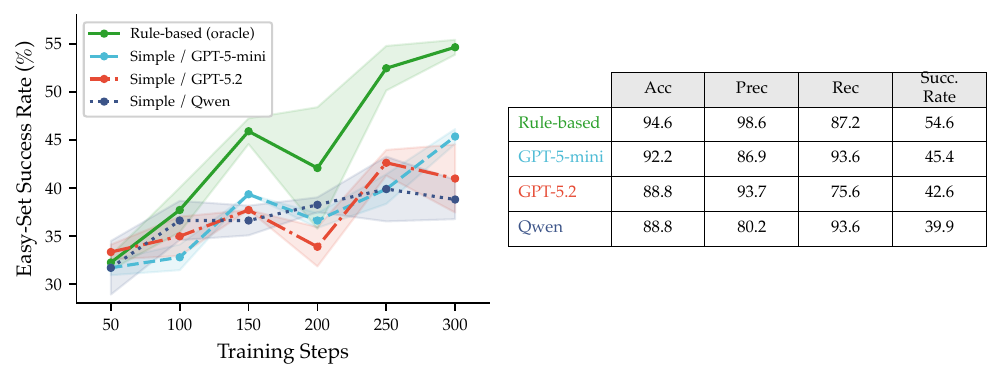}
    \vspace{-5mm}
    \caption{On-policy RL training curves with different judge reward signals (mean $\pm$ std over 3 seeds). Inset table: judge quality on the AndroidWorld subset (Acc/Prec/Rec \%) and best easy-set success rate. Training performance tracks judge accuracy (rule-based $>$ GPT-5-mini $>$ \{GPT-5.2, Qwen\}). The equal-accuracy pair is ordered by precision under the best-checkpoint convention.}
    \label{fig:training_curves}
    \end{figure*}

    Figure~\ref{fig:training_curves} shows that, in this setting, training outcomes follow judge accuracy: rule-based (94.6\% acc) $\to$ 54.6\% best easy-set success rate, GPT-5-mini (92.2\%) $\to$ 45.4\%, GPT-5.2 (88.8\%) $\to$ 42.6\%, Qwen (88.8\%) $\to$ 39.9\%. The same ordering holds on the complete 116-task suite (36.8\%, 30.2\%, 27.9\%, 26.4\%; Appendix~\ref{app:training}). GPT-5.2 and Qwen have identical accuracy but opposite precision--recall profiles, and the higher-precision GPT-5.2 (precision 93.7\% vs.\ 80.2\%) reaches a 2.7pp higher easy-set success rate, consistent with false positives directly rewarding incorrect behavior~\citep{huang2024darkside}. This comparison depends on the best-checkpoint convention: at the fixed final checkpoint the two conditions tie on the full task set (Appendix~\ref{app:training}). We therefore regard it as suggestive within our study.
    
    \subsection{Failure analysis}
    \label{sec:failure}
    
    Beyond aggregate metrics, we analyze cases where judges systematically fail to understand \emph{why} they fail and whether failure patterns depend on the LLM backbone.

    \begin{figure*}[t]
    \centering
    \includegraphics[width=0.90\textwidth]{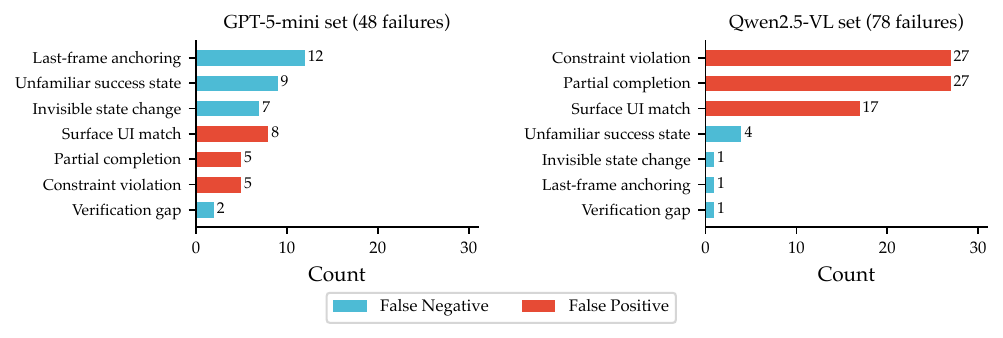}
    \vspace{-5mm}
    \caption{Root cause taxonomy for hard-core failures. The GPT set is dominated by false negatives (last-frame anchoring, unfamiliar success state), while the Qwen set is dominated by false positives (constraint violation, partial completion). Surface UI match is a shared weakness across both backbones.}
    \label{fig:failure_taxonomy}
    \end{figure*}

    \subsubsection{Method}
    
    For each of the 5 judge methods adapted from existing benchmarks (this analysis excludes our simple baseline), we identify trajectories where the judge prediction disagrees with human ground truth. We run this analysis separately for two LLM backbones, GPT-5-mini and Qwen2.5-VL-72B, to compare failure profiles. We define \emph{hard-core failures} as cases where at least 4 of the 5 methods err, indicating failures that are robust to judge design choices and likely reflect fundamental limitations. For each hard-core failure, we examine the task goal, ground-truth label, trajectory screenshots, and the reasoning produced by all 5 judges, then categorize the root cause into a taxonomy of error types (detailed in Appendix~\ref{app:failure}).
    
    \subsubsection{Opposite failure profiles across backbones}
    
    The failure profiles reveal a striking asymmetry. GPT-based judges produce 48 hard-core failures dominated by \emph{false negatives} (30~FN, 18~FP): they are too conservative, failing to recognize successful trajectories. Qwen-based judges produce 78 hard-core failures dominated by \emph{false positives} (71~FP, 7~FN): they are too permissive, accepting failed trajectories as successes. Only 22 cases appear in both sets (17 false positives and 5 false negatives shared by both backbones), suggesting that these cases are ambiguous regardless of the backbone.
    
    \subsubsection{Root cause taxonomy}
    
    Figure~\ref{fig:failure_taxonomy} shows the root cause categories for each backbone set, revealing qualitatively different failure modes. Representative examples for each category are in Appendix~\ref{app:failure_examples} (Figures~\ref{fig:failure_examples_fn}--\ref{fig:failure_examples_fp}).
    
    To assess the reliability of this classification, two raters independently categorized a random sample of 30 hard-core failure cases into the seven categories, agreeing on 29 of 30 (Cohen's $\kappa = 0.957$; Appendix~\ref{app:taxonomy_reliability}).

    \noindent{\bf GPT set (FN-dominated).} The top failure mode is \emph{last-frame anchoring} (12 cases): judges over-rely on the final screenshot and miss mid-trajectory completion evidence. \emph{Unfamiliar success state} (9 cases) and \emph{invisible state change} (7 cases) follow: judges either do not recognize valid success patterns or miss changes not visually evident in screenshots. Among FPs, \emph{surface UI match} (8 cases) leads: judges conclude success from superficial visual similarity without verifying the actual answer.
    
    \noindent{\bf Qwen set (FP-dominated).} \emph{Constraint violation} and \emph{partial completion} (27 cases each) dominate: judges miss task-specific constraints (\eg, ``nearest'' parking lot) or accept incomplete results as success. \emph{Surface UI match} (17 cases) is the third category. Representative examples are in Appendix~\ref{app:failure_examples}.
    
    \section{Discussion}
    \label{sec:discussion}
    
    \noindent{\bf Judges should be evaluated, not assumed reliable.}
    Judge accuracy ranges from 76\% to 91\% depending on method and backbone, and this variation shifts agent rankings by up to 13 positions and meaningfully affects training outcomes. Benchmarks should report judge reliability alongside agent performance.
    
    \noindent{\bf Elaborate judge pipelines do not consistently outperform a simple baseline.}
    Method choice, backbone choice, and their interaction all affect accuracy, but the added complexity of purpose-built methods is not reliably rewarded: the simple baseline is never significantly less accurate than a purpose-built method on any backbone, it uses a fraction of the tokens of the two most elaborate pipelines, and the weakest methods are purpose-built. Once those are excluded, backbone choice explains most of the remaining variance (Appendix~\ref{app:variance}). Among competitive methods, upgrading the backbone yields greater returns than engineering elaborate judge prompts.
    
    \noindent{\bf Different applications demand different judge profiles.}
    For \emph{evaluation}, F1 and balanced accuracy best predict reliability, while precision alone does not predict ranking fidelity. For \emph{training}, our matched-accuracy comparison suggests that false positives, which directly corrupt the reward signal~\citep{huang2024darkside}, are the more damaging error. These results suggest that balanced judges may be more suitable for leaderboards, while judges with strong false-positive control may be preferable as reward signals. Aggregating judges is a simple way to choose the operating point: with GPT-5-mini, accepting a trajectory when at least 3 of the 6 methods vote success gives 90.0\% precision and 93.3\% recall, whereas requiring 5 votes gives 95.1\% and 67.4\% (Appendix~\ref{app:thresholds}).
    
    \noindent{\bf Judge selection transfers across source benchmarks.}
    Selecting the best of the 30 judges on five source benchmarks and testing it on the sixth yields a judge ranked 1st to 9th of 30 on the held-out benchmark, on average 2.2pp below the best judge there (Appendix~\ref{app:lobo}).

    \noindent{\bf Failure modes are structurally addressable.}
    \emph{Last-frame anchoring} can be mitigated by providing more screenshots, consistent with our ablation. \emph{Surface UI match}, a shared weakness, motivates proactive verification approaches~\citep{vagen2026, prore2025} where judges interact with the environment.
    
    \section{Conclusion}
    \label{sec:conclusion}
    
    We introduced \bench{}, a benchmark for evaluating LLM-as-a-judge methods on mobile agent trajectories. Through 931 human-annotated trajectories spanning 6 benchmarks, 4 agents, and 59 apps, we systematically evaluated 6 judge methods across 5 LLM backbones. Our key findings are: (1)~a simple baseline judge with sampled screenshots is competitive with, and often exceeds, purpose-built methods, indicating that more elaborate judge pipelines do not consistently improve judge quality; (2)~benchmark quality metrics, particularly F1 and balanced accuracy, are strongly associated with agent ranking reliability on our benchmark, and in our training study the judge with stronger false-positive control reached a higher best-checkpoint success rate; and (3)~failure analysis reveals that different LLM backbones produce qualitatively opposite failure profiles, with surface UI match as a shared weakness. We hope that \bench{} encourages the community to treat judge evaluation as a first-class concern and provides a foundation for developing more reliable evaluation methods for mobile agents.
    \label{end:content}%

    \section*{Limitations}
    \label{sec:limitations}
    Our benchmark evaluates binary success/failure judgments. Finer-grained dimensions such as trajectory optimality, efficiency, and partial progress~\citep{lu2025agentrewardbench} are left to future work. Binary success is, however, the signal that the built-in checkers of current mobile-agent benchmarks produce and that mobile RL pipelines consume, which makes it the natural first target for a judge benchmark.

    Our training experiments are limited to the AndroidWorld environment and to a single judge method. The four reward sources are three LLM backbones of that judge plus a rule-based oracle. Training-time comparisons across judge methods are therefore narrower than the six-method offline study in \S\ref{sec:judge_eval}.

    The meta-correlation analysis is in-sample: judge quality and evaluation reliability are measured on the same 931 trajectories. Our transfer analysis holds out one of the six source benchmarks at a time (Appendix~\ref{app:lobo}).

    Finally, the root-cause classification in our failure analysis involves subjective judgment in borderline cases, although a double-coded sample of 30 cases shows high inter-rater agreement (Appendix~\ref{app:taxonomy_reliability}).

    \section*{Ethical Considerations}

    Judge unreliability is itself the primary risk this work concerns. LLM judges misclassify 9--24\% of mobile-agent trajectories (\S\ref{sec:judge_eval}). Treating their outputs as ground truth can distort leaderboards (\S\ref{sec:meta_corr}) and, when judges serve as reward signals, train agents on corrupted feedback
    (\S\ref{sec:training}). \bench{} is intended to mitigate this risk by quantifying it: we recommend reporting judge reliability alongside agent results and controlling false positives when judges are used as rewards.
    
    Documenting systematic judge weaknesses (\eg, surface UI match, \S\ref{sec:failure}) could in principle be used to game LLM judges. We consider the transparency benefit to outweigh this risk: identifying failure modes is a prerequisite for building robust, state-verifying judges, which our analysis motivates. Relatedly, agents trained against permissive judges may appear successful while violating task constraints. Our false-positive analysis (\S\ref{sec:training}) is aimed at preventing exactly this outcome in practice.
    
    All trajectories are collected in emulated Android environments. No real user data is involved and no agent action affects real services (Appendix~\ref{app:benchmark}). 

    \section*{Author Contributions}

    Ziqiang Wang led the research, including the construction of the benchmark, the experiments, the analysis, and the writing of the paper.
    Li Gu, Zhixiang Chi, Zhi Liu, and Seyed Mehdi Ayyoubzadeh contributed to the research discussions and to the revision of the paper.
    Yuanhao Yu served as the industrial project manager. Yang Wang served as the academic supervisor.

    \bibliography{mobilejudge}
    
    \newpage
    \appendix
    
    \section{Benchmark details}
    \label{app:benchmark}
    
    \paragraph{Unified environment.}
    We build a unified execution environment on top of the AndroidWorld codebase~\citep{rawles2025androidworld}, integrating apps and initialization logic from all 6 benchmarks. AndroidWorld tasks use a rooted system image without Google Play Store; other benchmarks use a separate system image with Google Play Store access (no root) and device backup/restore for task initialization. This reduces the maintenance burden of operating 6 separate codebases while preserving each benchmark's original task semantics.

    \paragraph{Licenses.}
    Of the assets we build on, the AndroidWorld codebase, B-MoCA, and A3 are released under the Apache-2.0 license, and SPA-Bench and AndroidLab under the MIT license, all permitting research use and redistribution. The AndroidArena repository specifies no license, and the AgentRewardBench code and dataset carry no standard license (the dataset is distributed under custom research terms of use, and we use no data from it): from AndroidArena we use task definitions, and from AgentRewardBench the judge-prompt design published in its paper, both for research evaluation with attribution. The trajectories, annotations, and evaluation code we release are our own artifacts.

    \paragraph{Release.}
    We release the 931 trajectories in a unified format (per-step screenshots, actions, UI elements, and agent reasoning), the task index, the final human success labels, and the evaluation code with the simple baseline judge. The data are distributed under CC~BY~4.0 and the code under the MIT license, with links given on the first page.

    \paragraph{Data provenance and privacy.}
    All trajectories are collected on emulated Android devices; no real user accounts or personal data appear in screenshots, UI trees, actions, or agent reasoning. Screen content is routine app UI from the benchmarks' controlled environments and contains no offensive material.

    \paragraph{Annotators.}
    The 9 annotators (\S\ref{sec:annotation}) are graduate students at the authors' institution, all proficient in English and experienced Android users. Annotation was conducted voluntarily as part of their funded research work, with the research use of the labels made clear; no additional task-specific payment was made. The study annotates machine-generated trajectories in emulated environments and involves no human-subjects data.

\begin{figure*}[t]
    \centering
    \includegraphics[width=0.32\textwidth]{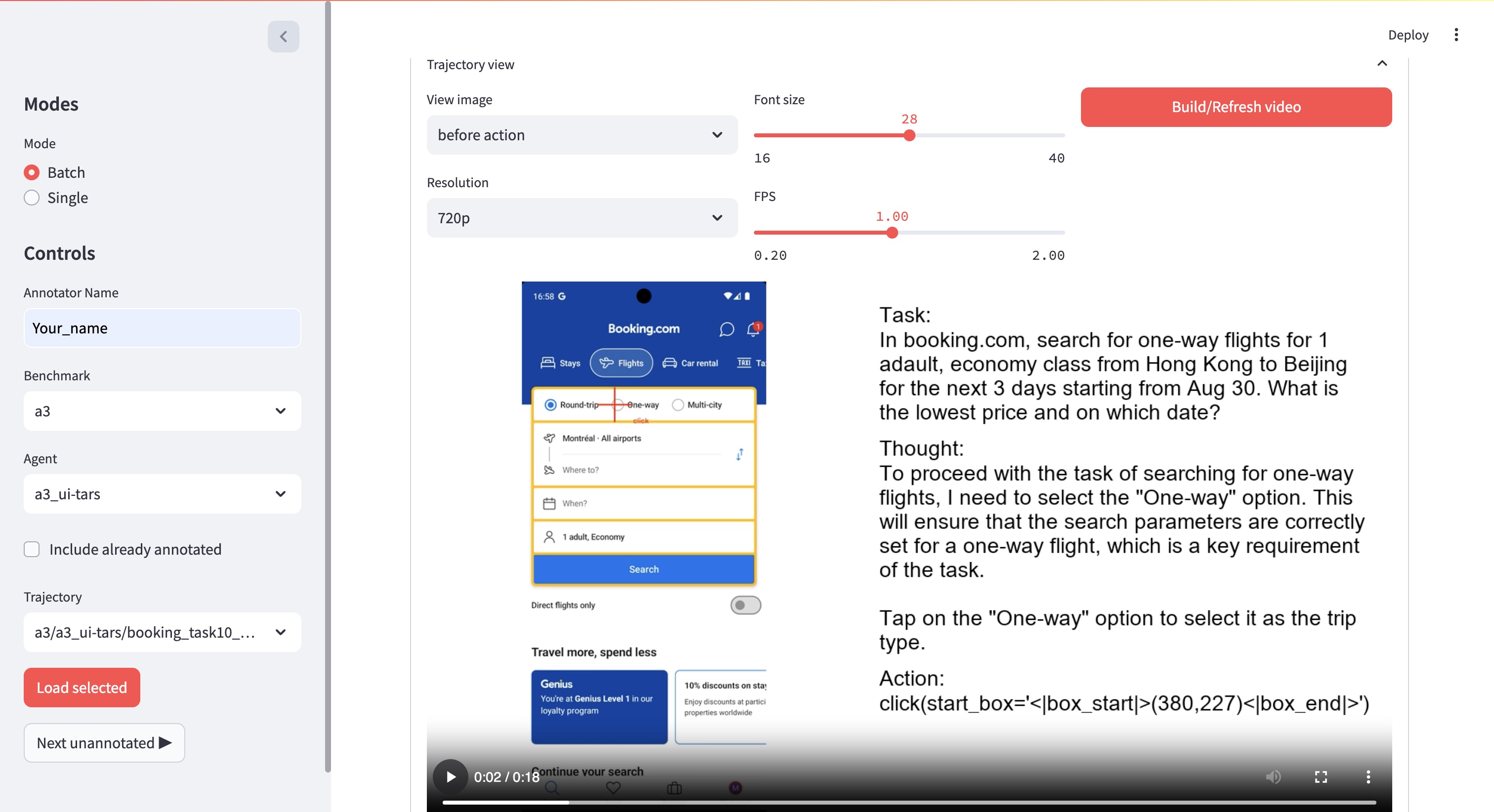}\hfill
    \includegraphics[width=0.32\textwidth]{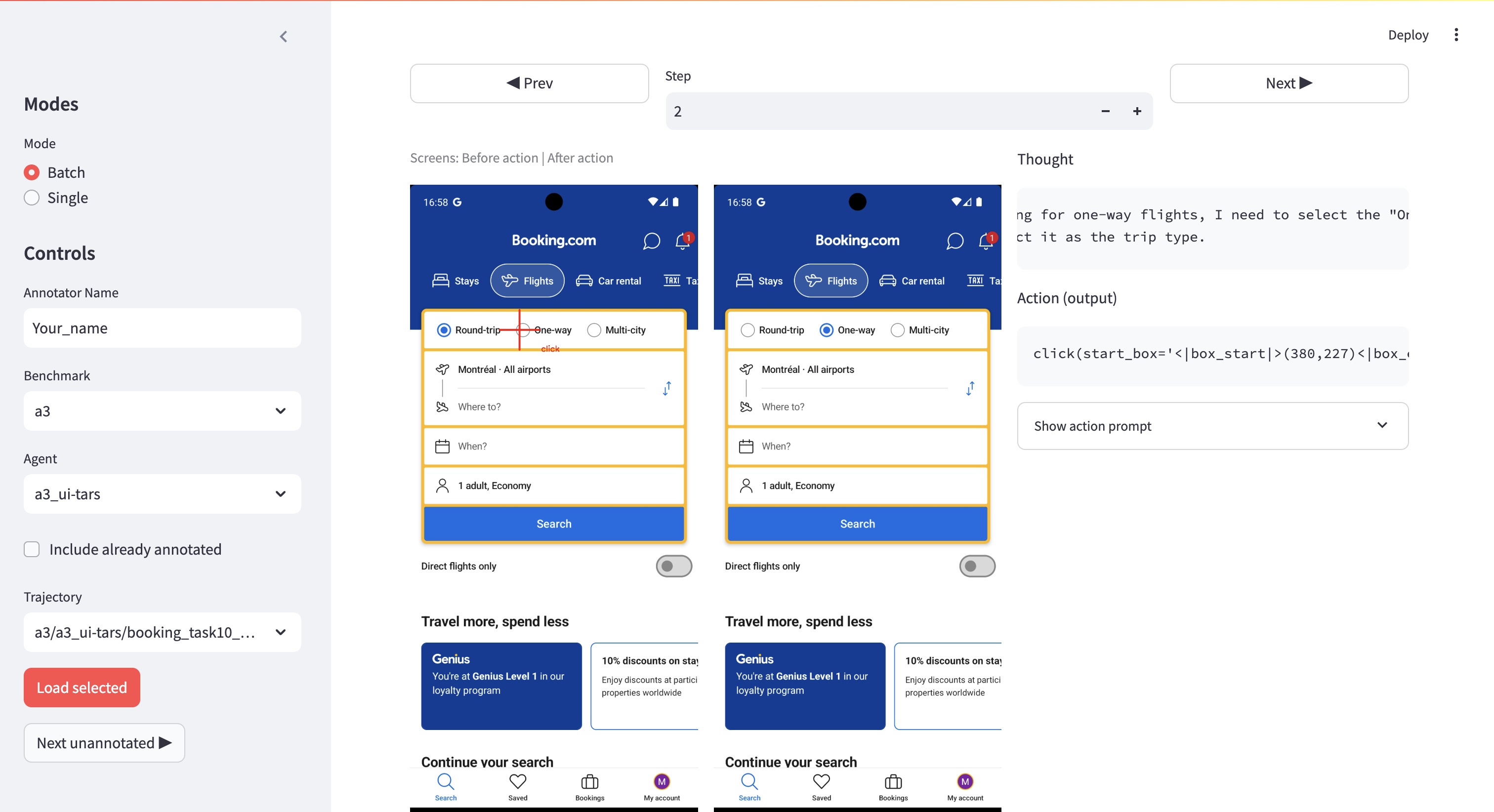}\hfill
    \includegraphics[width=0.32\textwidth]{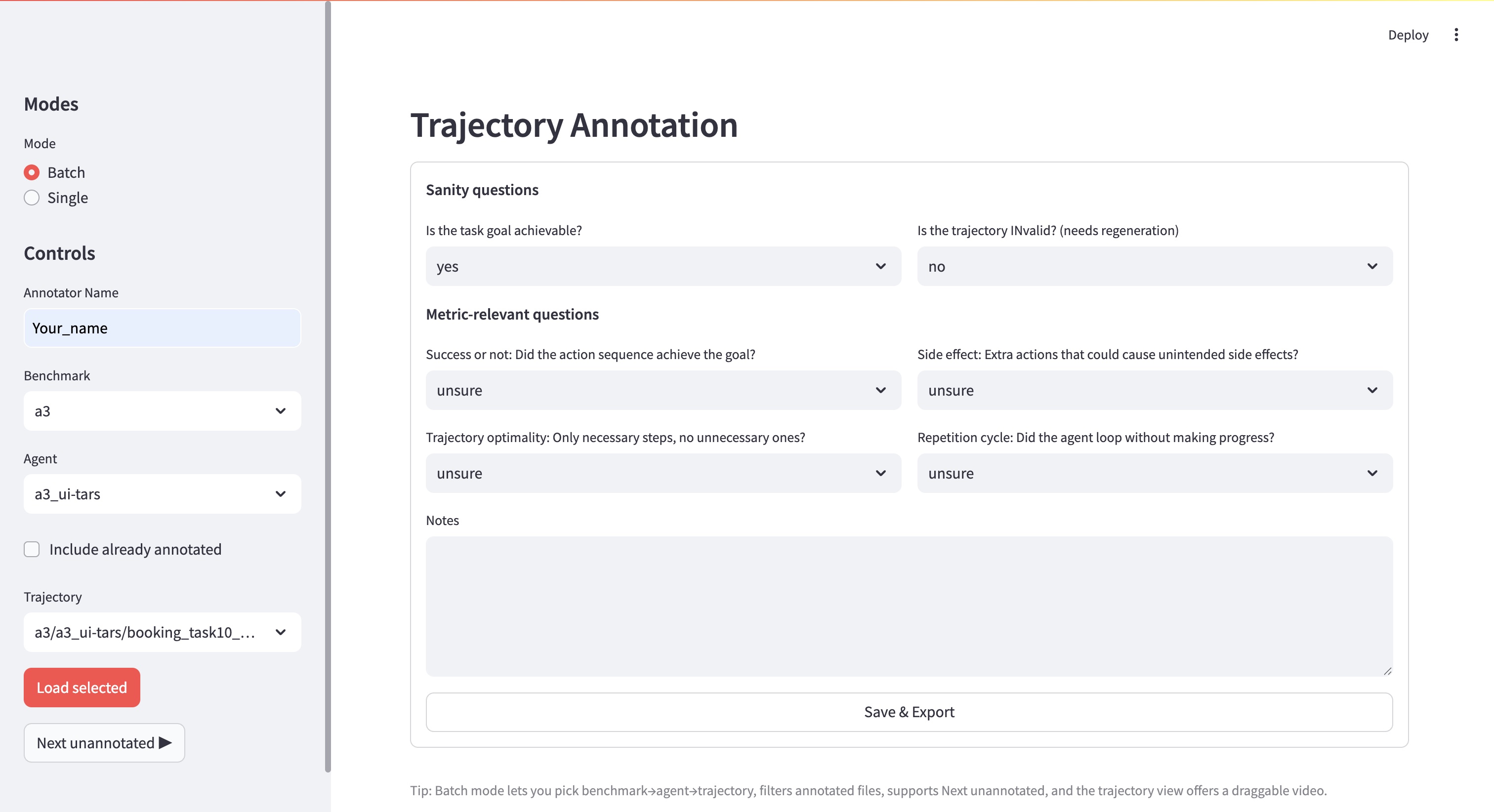}
    \caption{Annotation platform. Left: trajectory video view with task description, agent reasoning, and playback controls. Center: step-by-step viewer showing before/after screenshots and action details. Right: annotation form collecting task achievability, success, optimality, side effects, repetition cycles, and free-text notes.}
    \label{fig:annotation_platform}
    \end{figure*}

    \paragraph{Annotation platform.}
    We build a custom Streamlit-based annotation tool for trajectory-level labeling (Figure~\ref{fig:annotation_platform}). The tool supports two modes: \emph{batch mode}, where annotators navigate through benchmark$\to$agent$\to$trajectory hierarchies with automatic filtering of already-annotated items, and \emph{single mode} for ad-hoc trajectory inspection. For each trajectory, the platform provides: (1)~a \emph{trajectory video} synthesized from the screenshot sequence with configurable resolution, font size, and playback speed, enabling rapid overview of the full trajectory; (2)~a \emph{step viewer} showing before/after screenshots, the agent's chain-of-thought reasoning, and the executed action for each step; and (3)~an \emph{annotation form} collecting labels for task achievability, success, trajectory optimality, side effects, repetition cycles, and free-text notes. Annotations are exported as structured JSON files mirroring the source directory hierarchy.
    
    \section{Comparison with AgentRewardBench}
    \label{app:arb_comparison}

    AgentRewardBench~\citep{lu2025agentrewardbench} is the closest existing benchmark to ours: both evaluate LLM judges of agent trajectories against human annotations. The two benchmarks are complementary, differing in agent domain and in the validation questions they target. Table~\ref{tab:arb_comparison} provides a structured comparison; all AgentRewardBench entries are taken from \citet{lu2025agentrewardbench}.

    \begin{table*}[t]
    \centering
    \footnotesize
    \begin{tabular}{@{}p{2.1cm}p{6.1cm}p{6.1cm}@{}}
    \toprule
     & \textbf{AgentRewardBench}~\citep{lu2025agentrewardbench} & \textbf{\bench{} (ours)} \\
    \midrule
    Domain & Web browsing: WebArena, VisualWebArena, AssistantBench, WorkArena, WorkArena++ & Mobile (Android): SPA-Bench, AndroidWorld, A3, AndroidArena, B-MoCA, AndroidLab; 59 real apps \\
    \addlinespace
    Trajectory observations & Browser screenshots, DOM, and accessibility tree; judges evaluate the final browser state (screenshot or accessibility tree), for some judges together with the agent's thought and action sequence & Per-step (screenshot, action, UI tree, agent reasoning) tuples; judge methods span full-sequence and final-state designs, image and text inputs \\
    \addlinespace
    Judges evaluated & 3 judge designs (AER, NNetNav, and a simplified judge), instantiated as 12 judges over 4 LLM backbones & 6 judge methods (adapted from SPA-Bench, A3 with two modes, AndroidArena, and AgentRewardBench, plus a simple baseline), fully crossed with 5 LLM backbones: 30 judges \\
    \addlinespace
    Human annotation & Each trajectory reviewed by an expert (success, side effects, repetitiveness); 89.3\% success agreement measured on one agent--benchmark subset & Each trajectory independently labeled by 2--4 of 9 annotators; 88.4\% pairwise success agreement; disagreements resolved by discussion \\
    \addlinespace
    Judge validation & Per-trajectory precision, recall, and F1 against expert labels; precision emphasized as the primary metric, motivated by rejection fine-tuning and reward modeling; expert-vs-automatic success-rate comparison across agents & Per-trajectory accuracy, balanced accuracy, precision, recall, F1; meta-level validation linking judge metrics to agent-ranking fidelity and success-rate estimation error over 24 agents, with task-cluster bootstrap CIs \\
    \addlinespace
    Downstream RL & Motivates judge use for fine-tuning and RL rewards; no training experiment & On-policy GRPO training with judge-based rewards (4 reward configurations, 3 seeds), linking judge metrics to training outcomes \\
    \addlinespace
    Failure analysis & Four judge-error categories with examples & Seven-category root-cause taxonomy of hard-core failures, contrasted across two backbones; double-coded sample: $\kappa=0.957$ \\
    \bottomrule
    \end{tabular}
    \caption{Structured comparison of AgentRewardBench~\citep{lu2025agentrewardbench} and \bench{}. The two benchmarks cover different agent domains and are complementary; \bench{} additionally validates judge metrics against downstream evaluation and training utility.}
    \label{tab:arb_comparison}
    \end{table*}

    \section{Judge method details}
    \label{app:methods}
    
    The four existing judge methods (SPA-Bench, A3, AndroidArena, AgentRewardBench) use prompts adapted from their respective original papers. Below we present the prompt for our simple baseline judge, which is the only method we design from scratch.

    \paragraph{Decoding and backbone versions.}
    All judges decode greedily (temperature 0), except the A3 judges, whose calls use each provider's default sampling settings, and the GPT-5 family, whose API fixes sampling at the model default. Backbone version strings: \texttt{gpt-5-mini} and \texttt{gpt-5.2} via the OpenAI API; \texttt{google/gemini-3-flash-preview}, \texttt{z-ai/glm-4.6v}, and \texttt{anthropic/claude-sonnet-4.5} via OpenRouter; \texttt{Qwen2.5-VL-72B-Instruct} served from a local vLLM endpoint.
    
    \subsection{Simple baseline judge prompt}
    
    \noindent\textbf{System prompt:}
    \begin{quote}
    \small\ttfamily
    You are an expert evaluator for mobile agent task completion. Your role is to determine whether a given task has been successfully completed based on the provided trajectory evidence.
    
    \medskip
    \noindent Evaluation Guidelines:
    
    \medskip
    \noindent 1. \textbf{Outcome-Focused}: Judge based on whether the primary objective was achieved, not the path taken.
    
    \noindent 2. \textbf{Use All Provided Evidence}: Base your judgment on all available information --- screenshots, actions, agent reasoning, and UI element lists. Do not assume information beyond what is provided.
    
    \noindent 3. \textbf{Mid-Trajectory Success}: The task may be completed in an intermediate step rather than the final one. Additional actions after completion do not invalidate success, unless they explicitly undo it.
    
    \noindent 4. \textbf{Corrective Actions}: If the agent made mistakes but corrected them and achieved the goal, that counts as success.
    
    \noindent 5. \textbf{UI State Indicators}: Pay attention to visual indicators like selected tabs, checkboxes, highlighted items, and confirmation messages.
    
    \medskip
    \noindent Be balanced in your judgment --- avoid being overly strict (missing true successes) or overly lenient (accepting failures).
    \end{quote}
    
    \noindent\textbf{User prompt header:}
    \begin{quote}
    \small\ttfamily
    \#\# Task\\
    \{task\}
    
    \medskip
    \noindent \#\# Agent Trajectory\\
    The following shows the agent's execution trajectory in chronological order. Each step's screenshot is placed immediately after its description.
    \end{quote}
    
    \noindent\textbf{Step template} (with all optional components enabled):
    \begin{quote}
    \small\ttfamily
    \#\#\# Step \{step\_num\}\\
    \textbf{Action}: \{action\}\\
    \textbf{Agent Reasoning}: \{reasoning\}\\
    \textbf{UI Elements}: \{ui\_elements\}\\
    {[Screenshot attached]}
    \end{quote}
    
    \noindent\textbf{User prompt footer:}
    \begin{quote}
    \small\ttfamily
    \#\# Your Judgment\\
    Based on the trajectory above, determine if the task was successfully completed.
    
    \medskip
    \noindent Reply in the following format:\\
    Result: <1 for success, 0 for failure>\\
    Confidence: <high/medium/low>\\
    Reason: <brief explanation in 1--2 sentences>
    \end{quote}
    
    \noindent The step template varies depending on ablation configuration: agent reasoning and UI elements are independently toggled on/off. Screenshots are uniformly sampled from the trajectory (always including the first and last frames) and resized to a configurable resolution before encoding.
    
    \subsection{Uniform vs.\ event-based screenshot sampling}
    \label{app:sampling}

    A possible concern with uniform screenshot sampling is that it may miss transient evidence (\eg, a momentary confirmation toast). We therefore implemented an event-based sampler that prioritizes frames associated with informative actions (text entry, app switches, navigation, trajectory-terminating actions, and transitions between action types) and compared it with uniform sampling using the simple baseline judge (GPT-5-mini backbone) with all other settings identical. At the operating budget of 48 frames, only 1.3\% of trajectories (12/931) exceed the budget, so the two strategies select identical frames for the remaining 98.7\% by construction; on the 12 differing trajectories, the event-based sampler judges 11 correctly and the uniform sampler 10. Because so few trajectories are subsampled at 48 frames, we also ran a stress test at a tight 8-frame budget, where 61.9\% of trajectories are subsampled. Accuracy is 89.6\% (uniform) vs.\ 89.3\% (event-based); the two samplers disagree on 6.1\% of trajectories (57/931), and the disagreements split nearly evenly (30 uniform-correct vs.\ 27 event-correct). Uniform sampling therefore does not appear to limit judge quality at the trajectory lengths and input budgets we evaluate.

    \subsection{Final-frame baseline}
    \label{app:final_frame}

    We run the simple baseline prompt with only the final screenshot (max long edge 600px, about 1k tokens per trajectory), the same visual evidence that A3 final-state receives. Table~\ref{tab:final_frame} reports the result on four backbones. In paired task-cluster bootstrap comparisons, the final-frame baseline is significantly more accurate than A3 final-state with GPT-5-mini ($+7.3$pp, 95\% CI $[+4.3, +10.3]$) and GLM ($+4.4$pp $[+1.5, +7.3]$) and tied with it on Gemini ($+1.7$pp $[-0.3, +3.8]$) and Claude ($+2.0$pp $[-0.6, +4.7]$). AgentRewardBench, which adds the UI element list and a step summary to the final screenshot, is more accurate than the final-frame baseline on Gemini, GLM, and Claude and tied with it on GPT-5-mini. Going from one frame to the default 48 adds 2.4--7.0pp.

    \begin{table}[h]
    \centering
    \footnotesize
    \setlength{\tabcolsep}{3.2pt}
    \begin{tabular}{lccccccc}
    \toprule
     & \multicolumn{4}{c}{\textbf{Final frame}} & \multicolumn{3}{c}{\textbf{Reference acc.}} \\
    \cmidrule(lr){2-5}\cmidrule(lr){6-8}
    \textbf{Backbone} & \textbf{Acc} & \textbf{Prec} & \textbf{Rec} & \textbf{Tok.} & \textbf{A3-f} & \textbf{ARB} & \textbf{Base} \\
    \midrule
    GPT & 83.8 & 86.5 & 82.1 & 1.0 & 76.4 & 82.8 & 90.8 \\
    Gemini & 85.5 & 85.1 & 88.0 & 1.7 & 83.8 & 89.3 & 90.9 \\
    GLM & 82.7 & 81.1 & 87.6 & 1.2 & 78.1 & 86.1 & 84.9 \\
    Claude & 80.5 & 82.7 & 79.7 & 0.9 & 78.4 & 87.2 & 86.0 \\
    \bottomrule
    \end{tabular}
    \caption{Simple baseline with only the final screenshot (Final frame). Tok.: mean tokens per trajectory (thousands). Reference accuracies are from Table~\ref{tab:main_results}: A3 final-state (A3-f), AgentRewardBench (ARB), and the default 48-frame baseline (Base).}
    \label{tab:final_frame}
    \end{table}

    \subsection{WebJudge as an additional judge}
    \label{app:webjudge}

    WebJudge~\citep{xu2025webjudge} is a three-stage judge for web agent trajectories: it extracts key points from the task, scores each screenshot for relevance, and judges the outcome from the retained screenshots and the action history. We evaluate it in two forms (Table~\ref{tab:webjudge}). The first is the pipeline driven by GPT-5-mini, with the prompt wording adapted from web to mobile. It reaches 87.6\% accuracy. In paired bootstrap comparisons with the other GPT-5-mini judges, it is more accurate than AgentRewardBench and both A3 modes, tied with AndroidArena, and less accurate than SPA-Bench ($-3.0$pp, 95\% CI $[-5.2, -0.6]$) and the simple baseline ($-3.1$pp $[-5.4, -0.9]$). Because it scores every screenshot in a separate call, it uses 70.7k tokens per trajectory, 20$\times$ the baseline. The second is the released WebJudge-7B checkpoint, fine-tuned on web trajectories and run with its original prompts. It reaches 76.3\% accuracy, below all 30 judges of Table~\ref{tab:full_results}. Both results are consistent with the finding that more elaborate pipelines do not consistently improve judge quality.

    \begin{table}[h]
    \centering
    \footnotesize
    \setlength{\tabcolsep}{2.5pt}
    \begin{tabular}{lccccc}
    \toprule
    \textbf{Judge} & \textbf{Acc} & \textbf{Prec} & \textbf{Rec} & \textbf{F1} & \textbf{Tok.} \\
    \midrule
    WebJudge / GPT & 87.6 & 92.2 & 83.7 & 87.8 & 70.7 \\
    WebJudge-7B & 76.3 & 71.1 & 92.9 & 80.5 & 25.4 \\
    \midrule
    Simple Baseline / GPT & 90.8 & 88.3 & 95.1 & 91.6 & 3.5 \\
    \bottomrule
    \end{tabular}
    \caption{WebJudge on the 931 trajectories: the pipeline driven by GPT-5-mini and the released 7B checkpoint. Tok.: mean tokens per trajectory (thousands).}
    \label{tab:webjudge}
    \end{table}

    \section{Full evaluation results}
    \label{app:full_results}
    
    \subsection{Evaluation metric definitions}
    \label{app:metrics}
    
    Given a set of trajectories with human ground-truth labels and judge predictions, we compute the following metrics.
    
    \paragraph{Trajectory-level classification.}
    Let TP, FP, TN, FN denote the counts of true positives, false positives, true negatives, and false negatives, where a ``positive'' is a successful trajectory. We compute:
    \begin{itemize}
        \item $\mathrm{Accuracy} = (\mathrm{TP} + \mathrm{TN}) \,/\, (\mathrm{TP} + \mathrm{TN} + \mathrm{FP} + \mathrm{FN})$
        \item $\mathrm{Precision} = \mathrm{TP} \,/\, (\mathrm{TP} + \mathrm{FP})$
        \item $\mathrm{Recall} = \mathrm{TP} \,/\, (\mathrm{TP} + \mathrm{FN})$
        \item $\mathrm{F1} = 2 \cdot \mathrm{Precision} \cdot \mathrm{Recall} \,/\, (\mathrm{Precision} + \mathrm{Recall})$
        \item $\mathrm{Balanced\;Acc} = \frac{1}{2}\!\left(\frac{\mathrm{TP}}{\mathrm{TP}+\mathrm{FN}} + \frac{\mathrm{TN}}{\mathrm{TN}+\mathrm{FP}}\right)$
    \end{itemize}
    Precision measures how often the judge's success predictions are correct; recall measures how often truly successful trajectories are identified. Our dataset is approximately balanced (53\% positive), so accuracy and balanced accuracy are close, but we report both for completeness.
    
    \paragraph{Agent-level reliability.}
    For each agent $a$ (defined as a benchmark--model pair, \eg, \texttt{android\_world/gpt-5-mini}), we compute the success rate under human labels ($h_a$) and under judge predictions ($j_a$). We then measure reliability across $N{=}24$ agents via two complementary metrics:
    \begin{itemize}
        \item \textbf{Ranking fidelity}: Spearman rank correlation $\rho_s$ between the vectors $(h_1, \ldots, h_N)$ and $(j_1, \ldots, j_N)$. High $\rho_s$ means the judge preserves the relative ordering of agents.
        \item \textbf{Rate estimation error}: Mean absolute error $\mathrm{MAE} = \frac{1}{N}\sum_{a}|h_a - j_a|$. Low MAE means the judge accurately estimates absolute success rates per agent.
    \end{itemize}
    
    \paragraph{Meta-correlation.}
    To assess whether benchmark quality predicts downstream evaluation reliability, we compute Spearman $\rho_s$ between a quality metric vector (one value per judge variant, \eg, accuracy) and a reliability metric vector (one value per judge variant, \eg, agent-level $\rho_s$), across all 30 judge variants.
    
    \subsection{Agent-level data points}
    \label{app:agents}
    
    Table~\ref{tab:agent_list} lists the 24 agents (benchmark--model combinations) used for computing ranking fidelity and rate estimation error. SPA-Bench single-app and cross-app trajectories are merged into one agent per model.
    
    \begin{table}[h]
    \centering
    \small
    \setlength{\tabcolsep}{3.5pt}  %
    \begin{tabular}{llrr}
    \toprule
    \textbf{Agent} & \textbf{Model} & \textbf{Traj.} & \textbf{Rate} \\
    \midrule
    arena / gpt-5-mini         & gpt-5-mini &  44 & 0.886 \\
    bmoca / gpt-5-mini         & gpt-5-mini &  31 & 0.806 \\
    a3 / gpt-5-mini            & gpt-5-mini &  45 & 0.800 \\
    android-lab / gpt-5-mini   & gpt-5-mini &  20 & 0.800 \\
    spa / gpt-5-mini           & gpt-5-mini &  67 & 0.761 \\
    android\_world / gpt-5-mini & gpt-5-mini &  52 & 0.692 \\
    \midrule
    arena / ui-tars             & ui-tars    &  41 & 0.683 \\
    a3 / ui-tars               & ui-tars    &  40 & 0.675 \\
    bmoca / ui-tars             & ui-tars    &  31 & 0.516 \\
    android-lab / ui-tars       & ui-tars    &  20 & 0.500 \\
    spa / ui-tars               & ui-tars    &  53 & 0.491 \\
    android\_world / ui-tars    & ui-tars    &  50 & 0.340 \\
    \midrule
    arena / llama               & llama      &  34 & 0.618 \\
    bmoca / llama               & llama      &  28 & 0.500 \\
    a3 / llama                  & llama      &  28 & 0.357 \\
    android\_world / llama      & llama      &  52 & 0.346 \\
    spa / llama                 & llama      &  51 & 0.294 \\
    android-lab / llama         & llama      &  19 & 0.263 \\
    \midrule
    a3 / qwen                   & qwen       &  41 & 0.537 \\
    arena / qwen                & qwen       &  32 & 0.531 \\
    bmoca / qwen                & qwen       &  33 & 0.515 \\
    android-lab / qwen          & qwen       &  19 & 0.368 \\
    spa / qwen                  & qwen       &  49 & 0.245 \\
    android\_world / qwen       & qwen       &  51 & 0.137 \\
    \bottomrule
    \end{tabular}
    \caption{The 24 agent-level data points used for computing ranking fidelity and rate estimation error. Rate is the human-annotated success rate. Agents are grouped by model and sorted by success rate within each group.}
    \label{tab:agent_list}
    \end{table}
    
    \subsection{Full classification results}
    \label{app:full_class}
    
    Table~\ref{tab:full_results} presents the complete evaluation results for all 30 judge variants, including precision, recall, F1, balanced accuracy, and agent-level reliability metrics.
    
    \begin{table*}[h]
    \centering
    \scriptsize
    \setlength{\tabcolsep}{3.5pt}
    \begin{tabular}{llccccccc}
    \toprule
    \textbf{Judge Method} & \textbf{Backbone} & \textbf{Acc} & \textbf{Prec} & \textbf{Rec} & \textbf{F1} & \textbf{BAcc} & \textbf{Ag.$\rho$} & \textbf{Ag.MAE} \\
    \midrule
    Baseline        & Gemini & 90.9 & 87.6 & 96.3 & 91.8 & 90.5 & 0.97 & 0.064 \\
    Baseline        & GPT    & 90.8 & 88.3 & 95.1 & 91.6 & 90.5 & 0.96 & 0.054 \\
    SPA-Bench       & GPT    & 90.6 & 87.8 & 95.5 & 91.5 & 90.3 & 0.97 & 0.054 \\
    SPA-Bench       & Gemini & 89.8 & 88.5 & 92.9 & 90.7 & 89.6 & 0.96 & 0.054 \\
    AgentRewardBench& Gemini & 89.3 & 88.6 & 91.7 & 90.1 & 89.2 & 0.97 & 0.040 \\
    AndroidArena    & GPT    & 88.2 & 89.0 & 88.6 & 88.8 & 88.1 & 0.95 & 0.049 \\
    AndroidArena    & Gemini & 87.7 & 88.4 & 88.4 & 88.4 & 87.7 & 0.94 & 0.050 \\
    AgentRewardBench& Claude & 87.2 & 88.3 & 87.4 & 87.8 & 87.2 & 0.97 & 0.042 \\
    AgentRewardBench& GLM    & 86.1 & 82.0 & 94.5 & 87.8 & 85.6 & 0.95 & 0.086 \\
    Baseline        & Claude & 86.0 & 87.4 & 86.0 & 86.7 & 86.0 & 0.95 & 0.052 \\
    Baseline        & Qwen   & 85.7 & 82.6 & 92.5 & 87.2 & 85.3 & 0.96 & 0.066 \\
    Baseline        & GLM    & 84.9 & 81.0 & 93.3 & 86.7 & 84.5 & 0.92 & 0.083 \\
    AgentRewardBench& Qwen   & 84.9 & 79.9 & 95.3 & 86.9 & 84.2 & 0.96 & 0.104 \\
    SPA-Bench       & Claude & 84.7 & 91.5 & 78.5 & 84.5 & 85.1 & 0.92 & 0.076 \\
    AndroidArena    & Qwen   & 84.6 & 81.8 & 91.3 & 86.3 & 84.1 & 0.84 & 0.083 \\
    AndroidArena    & GLM    & 83.9 & 79.7 & 93.5 & 86.1 & 83.3 & 0.92 & 0.100 \\
    A3 (final)      & Gemini & 83.8 & 84.5 & 84.7 & 84.6 & 83.7 & 0.94 & 0.059 \\
    A3 (essential)  & Gemini & 83.7 & 88.5 & 79.5 & 83.7 & 83.9 & 0.91 & 0.082 \\
    AgentRewardBench& GPT    & 82.8 & 90.7 & 75.2 & 82.2 & 83.3 & 0.90 & 0.086 \\
    SPA-Bench       & GLM    & 82.2 & 82.0 & 85.0 & 83.4 & 82.0 & 0.92 & 0.073 \\
    A3 (essential)  & GLM    & 81.9 & 85.2 & 79.7 & 82.3 & 82.1 & 0.91 & 0.081 \\
    A3 (final)      & Qwen   & 81.5 & 82.3 & 82.9 & 82.6 & 81.4 & 0.90 & 0.075 \\
    A3 (essential)  & Claude & 81.3 & 83.0 & 81.3 & 82.1 & 81.3 & 0.87 & 0.078 \\
    SPA-Bench       & Qwen   & 80.1 & 73.3 & 98.2 & 83.9 & 79.0 & 0.92 & 0.173 \\
    A3 (essential)  & Qwen   & 80.0 & 80.5 & 82.1 & 81.3 & 79.9 & 0.83 & 0.086 \\
    AndroidArena    & Claude & 79.0 & 91.1 & 66.9 & 77.1 & 79.8 & 0.69 & 0.148 \\
    A3 (final)      & Claude & 78.4 & 87.8 & 68.7 & 77.1 & 79.0 & 0.87 & 0.113 \\
    A3 (final)      & GLM    & 78.1 & 85.5 & 70.5 & 77.3 & 78.5 & 0.87 & 0.098 \\
    A3 (essential)  & GPT    & 78.1 & 92.3 & 63.8 & 75.5 & 78.9 & 0.90 & 0.159 \\
    A3 (final)      & GPT    & 76.4 & 90.5 & 61.9 & 73.5 & 77.3 & 0.77 & 0.142 \\
    \bottomrule
    \end{tabular}
    \caption{Full evaluation results for all 30 judge variants, sorted by accuracy. Acc/Prec/Rec/F1/BAcc are trajectory-level percentages. Ag.$\rho$ is the agent-level Spearman rank correlation with human rankings. Ag.MAE is the mean absolute error of per-agent success rate estimates.}
    \label{tab:full_results}
    \end{table*}

    \subsection{Uncertainty quantification and variance decomposition}
    \label{app:variance}

\begin{figure*}[h]
    \centering
    \includegraphics[width=\textwidth]{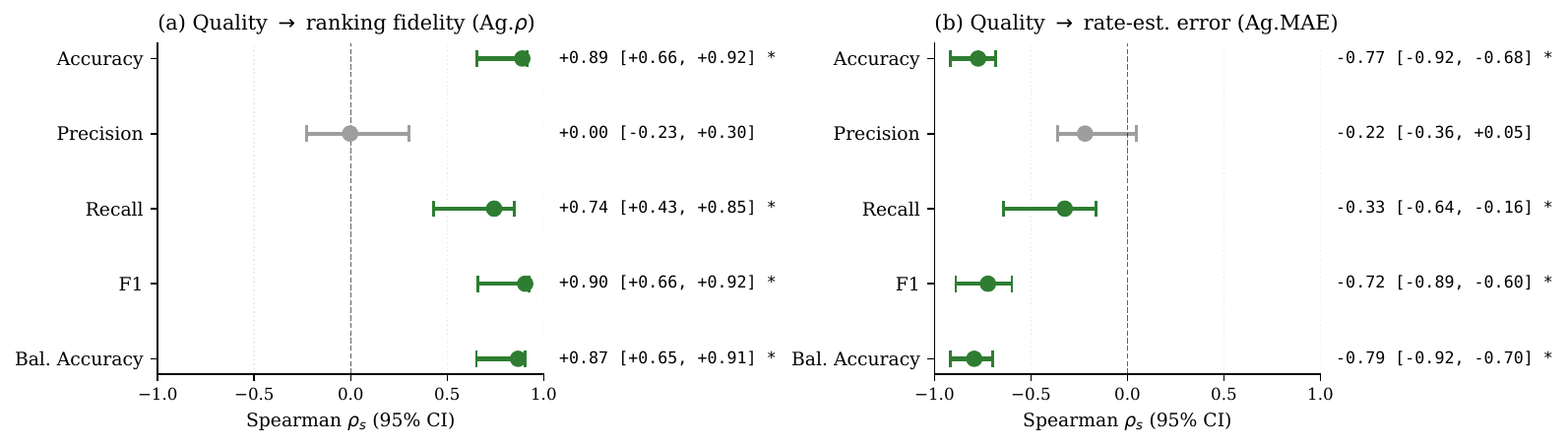}
    \caption{Task-cluster bootstrap 95\% CIs for the meta-correlation across the 30 judge variants. Green: interval excludes zero; grey: interval spans zero.}
    \label{fig:bootstrap_forest}
    \end{figure*}

    \paragraph{Task-cluster bootstrap.}
    The 95\% CIs in Table~\ref{tab:meta_corr} treat the 30 judge variants as fixed and quantify uncertainty from the benchmark sample. Each replicate resamples tasks with replacement within each of the 6 source benchmarks, where a sampled task carries all of its 1--4 agent trajectories (289 tasks and 931 trajectories in the original sample). We then recompute every judge's quality metrics, the per-agent reliability metrics, and the cross-judge meta-correlation (2{,}000 replicates; percentile intervals). Figure~\ref{fig:bootstrap_forest} visualizes the intervals. The conclusions are robust to leaving out any single backbone or judge method: F1 vs.\ ranking fidelity stays within $[0.87, 0.95]$ across all leave-one-out configurations, and precision remains non-predictive of ranking fidelity ($\rho_s \in [-0.15, 0.13]$). To measure provider-side nondeterminism, we re-ran the simple baseline judge three times on a stratified 100-trajectory subset with GPT-5-mini and Gemini~3~Flash. Accuracy varied across the three runs with a standard deviation of 2.1pp and 1.0pp, respectively, and the verdict changed on 6 and 4 of the 100 trajectories.

    \paragraph{Method vs.\ backbone variance decomposition.}
    Table~\ref{tab:variance_decomp} decomposes the total sum of squares of the 6$\times$5 method$\times$backbone quality grid into the method main effect, the backbone main effect, and the residual (interaction), with task-cluster bootstrap CIs. On the full grid the method share dominates, but it is driven by the two A3 modes, the weakest methods. Excluding these two (leaving AndroidArena, SPA-Bench, AgentRewardBench, and the baseline) reverses the accuracy shares to 0.11 (method) vs.\ 0.49 (backbone). Because this subset is selected by performance, we report the exclusion as a sensitivity analysis.

    \begin{table}[h]
    \centering
    \footnotesize
    \setlength{\tabcolsep}{2pt}
    \begin{tabular}{lccc}
    \toprule
    \textbf{Metric} & \textbf{Method} & \textbf{Backbone} & \textbf{Residual} \\
    \midrule
    Accuracy      & 0.49 {\scriptsize$[0.37, 0.59]$} & 0.21 {\scriptsize$[0.13, 0.30]$} & 0.30 {\scriptsize$[0.25, 0.39]$} \\
    F1            & 0.52 {\scriptsize$[0.43, 0.60]$} & 0.16 {\scriptsize$[0.11, 0.22]$} & 0.32 {\scriptsize$[0.27, 0.40]$} \\
    Bal.\ Acc     & 0.46 {\scriptsize$[0.33, 0.56]$} & 0.24 {\scriptsize$[0.15, 0.34]$} & 0.30 {\scriptsize$[0.24, 0.39]$} \\
    \midrule
    \multicolumn{4}{@{}l}{\emph{Competitive 4$\times$5 subgrid (excluding the two A3 modes)}} \\
    Accuracy      & 0.11 {\scriptsize$[0.04, 0.20]$} & 0.49 {\scriptsize$[0.34, 0.62]$} & 0.40 {\scriptsize$[0.29, 0.54]$} \\
    F1            & 0.12 {\scriptsize$[0.06, 0.21]$} & 0.38 {\scriptsize$[0.26, 0.51]$} & 0.49 {\scriptsize$[0.38, 0.61]$} \\
    Bal.\ Acc     & 0.10 {\scriptsize$[0.04, 0.19]$} & 0.54 {\scriptsize$[0.39, 0.66]$} & 0.37 {\scriptsize$[0.26, 0.50]$} \\
    \bottomrule
    \end{tabular}
    \caption{Share of total variance ($\eta^2$) across the 6$\times$5 judge grid attributable to the method and backbone main effects and the residual (interaction), with task-cluster bootstrap 95\% CIs. The lower block repeats the decomposition on the competitive 4$\times$5 subgrid that excludes the two A3 modes. The same task draws are used for both grids.}
    \label{tab:variance_decomp}
    \end{table}

    \subsection{Per-judge confidence intervals and paired comparisons}
    \label{app:paired}

    Table~\ref{tab:judge_ci} reports the accuracy of each of the 30 judges with a task-cluster bootstrap 95\% CI (tasks resampled with replacement within each source benchmark, 2{,}000 replicates). Table~\ref{tab:paired_diffs} reports the paired accuracy difference between the simple baseline and each purpose-built method on the same backbone, computed in every replicate on the trajectories that both judges evaluated. The column leader and the runner-up of Table~\ref{tab:main_results} are tied on every backbone: the baseline leads SPA-Bench by $+0.1$pp $[-1.8, +1.9]$ with GPT-5-mini and by $+1.1$pp $[-0.8, +2.8]$ with Gemini, and AgentRewardBench by $+0.9$pp $[-1.4, +3.1]$ with Qwen. AgentRewardBench leads the baseline by $+1.2$pp with GLM ($[-0.6, +3.1]$) and with Claude ($[-1.2, +3.5]$). Across the 25 comparisons, the baseline is significantly more accurate in 15 (all five against A3 final-state, four against A3 essential-states, three against AndroidArena, two against SPA-Bench, and one against AgentRewardBench) and significantly less accurate in none.

    \begin{table*}[h]
    \centering
    \footnotesize
    \setlength{\tabcolsep}{4pt}
    \begin{tabular}{lccccc}
    \toprule
    \textbf{Judge Method} & \textbf{Qwen} & \textbf{GPT} & \textbf{Gemini} & \textbf{GLM} & \textbf{Claude} \\
    \midrule
    AndroidArena & 84.6 {\scriptsize$[81.9, 87.0]$} & 88.2 {\scriptsize$[85.7, 90.3]$} & 87.7 {\scriptsize$[85.4, 89.8]$} & 83.9 {\scriptsize$[81.2, 86.5]$} & 79.0 {\scriptsize$[76.0, 81.9]$} \\
    A3 (final state) & 81.5 {\scriptsize$[78.4, 84.3]$} & 76.4 {\scriptsize$[73.2, 79.5]$} & 83.8 {\scriptsize$[81.0, 86.4]$} & 78.1 {\scriptsize$[74.8, 81.0]$} & 78.4 {\scriptsize$[75.4, 81.2]$} \\
    A3 (essential) & 80.0 {\scriptsize$[77.1, 82.7]$} & 78.1 {\scriptsize$[74.9, 81.0]$} & 83.7 {\scriptsize$[80.9, 86.2]$} & 81.9 {\scriptsize$[79.3, 84.4]$} & 81.3 {\scriptsize$[78.2, 84.1]$} \\
    SPA-Bench & 80.1 {\scriptsize$[77.0, 82.9]$} & 90.6 {\scriptsize$[88.4, 92.6]$} & 89.8 {\scriptsize$[87.5, 92.0]$} & 82.2 {\scriptsize$[79.4, 84.8]$} & 84.7 {\scriptsize$[82.1, 87.3]$} \\
    AgentRewardBench & 84.9 {\scriptsize$[82.3, 87.4]$} & 82.8 {\scriptsize$[79.7, 85.6]$} & 89.3 {\scriptsize$[87.1, 91.4]$} & 86.1 {\scriptsize$[83.6, 88.6]$} & 87.2 {\scriptsize$[84.7, 89.5]$} \\
    \midrule
    Simple Baseline & 85.7 {\scriptsize$[83.1, 88.3]$} & 90.8 {\scriptsize$[88.5, 92.8]$} & 90.9 {\scriptsize$[88.7, 92.9]$} & 84.9 {\scriptsize$[82.4, 87.4]$} & 86.0 {\scriptsize$[83.5, 88.4]$} \\
    \bottomrule
    \end{tabular}
    \caption{Judge accuracy (\%) with task-cluster bootstrap 95\% CIs (tasks resampled within each source benchmark; 2{,}000 replicates).}
    \label{tab:judge_ci}
    \end{table*}

    \begin{table*}[h]
    \centering
    \footnotesize
    \setlength{\tabcolsep}{4pt}
    \begin{tabular}{lccccc}
    \toprule
    \textbf{Baseline minus} & \textbf{Qwen} & \textbf{GPT} & \textbf{Gemini} & \textbf{GLM} & \textbf{Claude} \\
    \midrule
    AndroidArena & +1.1 {\scriptsize$[-1.1, +3.2]$} & +2.6$^{*}$ {\scriptsize$[+0.5, +4.6]$} & +3.1$^{*}$ {\scriptsize$[+1.3, +4.9]$} & +1.1 {\scriptsize$[-1.1, +3.3]$} & +7.0$^{*}$ {\scriptsize$[+4.1, +9.9]$} \\
    A3 (final state) & +4.2$^{*}$ {\scriptsize$[+1.6, +6.8]$} & +14.3$^{*}$ {\scriptsize$[+11.0, +17.6]$} & +7.1$^{*}$ {\scriptsize$[+4.7, +9.6]$} & +6.8$^{*}$ {\scriptsize$[+3.7, +9.8]$} & +7.6$^{*}$ {\scriptsize$[+4.6, +10.7]$} \\
    A3 (essential) & +5.7$^{*}$ {\scriptsize$[+2.7, +8.9]$} & +12.7$^{*}$ {\scriptsize$[+9.4, +15.9]$} & +7.2$^{*}$ {\scriptsize$[+4.5, +9.9]$} & +2.9 {\scriptsize$[+0.0, +5.8]$} & +4.7$^{*}$ {\scriptsize$[+1.9, +7.5]$} \\
    SPA-Bench & +5.6$^{*}$ {\scriptsize$[+3.2, +8.0]$} & +0.1 {\scriptsize$[-1.8, +1.9]$} & +1.1 {\scriptsize$[-0.8, +2.8]$} & +2.7$^{*}$ {\scriptsize$[+0.1, +5.1]$} & +1.3 {\scriptsize$[-1.2, +3.7]$} \\
    AgentRewardBench & +0.9 {\scriptsize$[-1.4, +3.1]$} & +8.0$^{*}$ {\scriptsize$[+5.2, +10.7]$} & +1.5 {\scriptsize$[-0.2, +3.2]$} & $-$1.2 {\scriptsize$[-3.1, +0.6]$} & $-$1.2 {\scriptsize$[-3.5, +1.2]$} \\
    \bottomrule
    \end{tabular}
    \caption{Paired accuracy difference (pp) between the simple baseline and each purpose-built method on the same backbone, with task-cluster bootstrap 95\% CIs. $^{*}$: the interval excludes zero.}
    \label{tab:paired_diffs}
    \end{table*}

    \subsection{Per-benchmark results}
    \label{app:per_benchmark}

\begin{figure*}[h]
    \centering
    \includegraphics[width=0.82\textwidth]{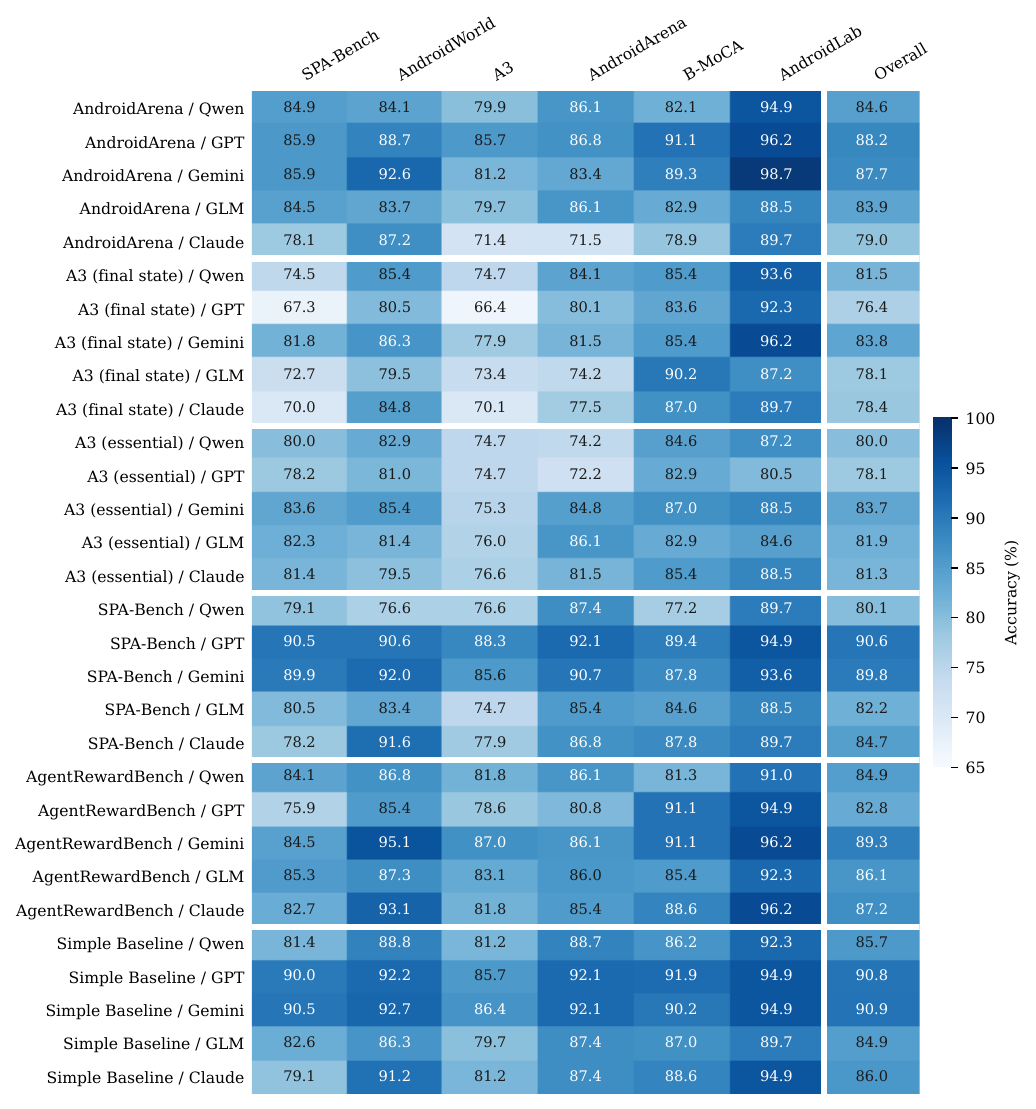}
    \caption{Accuracy (\%) of the 30 judges on each source benchmark and on the pooled 931 trajectories.}
    \label{fig:per_benchmark}
    \end{figure*}

    Figure~\ref{fig:per_benchmark} breaks the accuracy of the 30 judges down by source benchmark. The spread across judges on a single benchmark is 15--23pp, largest on SPA-Bench, A3, and AndroidArena and smallest on B-MoCA. The findings of \S\ref{sec:judge_eval} hold within each benchmark: for every backbone, the simple baseline is within 3.8pp of the best method on every benchmark (within 2.6pp for GPT-5-mini), AgentRewardBench leads on most benchmarks for GLM and Claude, and one of the two A3 modes is the weakest method in most backbone--benchmark cells. The clearest dataset-specific effect is on AndroidLab, whose trajectories are short and text-rich: the text-based AndroidArena judge is the strongest method there for the Qwen, GPT-5-mini, and Gemini backbones.

    \subsection{Leave-one-benchmark-out judge selection}
    \label{app:lobo}

    The six source benchmarks were built independently, with different task authors, apps, and evaluation paradigms. We use this to test whether a judge selected on the benchmark remains a good choice on data from a source it was not selected on. For each source benchmark, we select the most accurate of the 30 judges on the other five and evaluate it on the held-out one (Table~\ref{tab:lobo}). The selected judge ranks between 1st and 9th of 30 on the held-out benchmark, with a mean gap of 2.2pp to the best judge there (at most 4.5pp). At the level of method choice, selecting the best of the six methods within a backbone on five benchmarks yields a method that is in the top two on the held-out benchmark in 23 of the 30 backbone--benchmark folds (mean gap 1.0pp, at most 4.9pp). The selection is stable: the simple baseline in every fold for Qwen and Gemini and in four of six folds for GPT-5-mini (SPA-Bench in the other two), and AgentRewardBench in every fold for GLM and Claude. Agreement of the 30-judge ranking across benchmarks is moderate to high (Kendall's $W = 0.64$, and Spearman $\rho$ between each benchmark and the pooled remaining five ranges from 0.54 to 0.93).

    \begin{table*}[h]
    \centering
    \footnotesize
    \setlength{\tabcolsep}{3pt}
    \begin{tabular}{lrlcclcc}
    \toprule
    \textbf{Held-out benchmark} & \textbf{Traj.} & \textbf{Judge selected on the other five} & \textbf{Acc} & \textbf{Rank} & \textbf{Best judge on held-out} & \textbf{Acc} & \textbf{Gap} \\
    \midrule
    SPA-Bench & 220 & Baseline / GPT & 90.0 & 3 & SPA-Bench / GPT & 90.5 & 0.5 \\
    AndroidWorld & 205 & SPA-Bench / GPT & 90.6 & 9 & AgentRewardBench / Gemini & 95.1 & 4.5 \\
    A3 & 154 & Baseline / GPT & 85.7 & 4 & SPA-Bench / GPT & 88.3 & 2.6 \\
    AndroidArena & 151 & Baseline / Gemini & 92.1 & 1 & SPA-Bench / GPT & 92.1 & 0.0 \\
    B-MoCA & 123 & Baseline / Gemini & 90.2 & 5 & Baseline / GPT & 91.9 & 1.6 \\
    AndroidLab & 78 & Baseline / Gemini & 94.9 & 6 & AndroidArena / Gemini & 98.7 & 3.8 \\
    \bottomrule
    \end{tabular}
    \caption{Leave-one-benchmark-out selection among the 30 judges. Acc and Rank (of 30) are measured on the held-out benchmark; Gap is the accuracy difference (pp) to the best judge on that benchmark.}
    \label{tab:lobo}
    \end{table*}

    \subsection{Inference cost and latency}
    \label{app:cost}

    Table~\ref{tab:cost} reports, for each of the 30 judges, the mean number of tokens per trajectory (prompt plus completion, summed over all LLM calls the method makes) and the median wall-clock time per trajectory. The simple baseline uses 3.4--3.7k tokens on four backbones, in the same range as the final-state judges, and 14.3k on Gemini, whose image tokenization differs. SPA-Bench and A3 essential-states use several times more. Latency is indicative only: it includes provider queueing, Qwen is served locally while the other backbones are called through APIs, and the runs were not executed under identical concurrency.

    \begin{table}[h]
    \centering
    \footnotesize
    \setlength{\tabcolsep}{2.5pt}
    \begin{tabular}{lccccc}
    \toprule
    \textbf{Judge Method} & \textbf{Qwen} & \textbf{GPT} & \textbf{Gemini} & \textbf{GLM} & \textbf{Claude} \\
    \midrule
    \multicolumn{6}{@{}l}{\emph{Mean tokens per trajectory (thousands)}} \\
    AndroidArena & 11.9 & 8.5 & 6.2 & 12.6 & 7.4 \\
    A3 (final state) & 3.9 & 2.6 & 1.6 & 4.2 & 2.1 \\
    A3 (essential) & 63.8 & 18.8 & 13.3 & 38.6 & 16.2 \\
    SPA-Bench & 42.1 & 21.9 & 12.9 & 42.2 & 17.4 \\
    AgentRewardBench & 5.3 & 3.7 & 3.1 & 6.0 & 3.8 \\
    Simple Baseline & 3.4 & 3.5 & 14.3 & 3.7 & 3.6 \\
    \midrule
    \multicolumn{6}{@{}l}{\emph{Median latency per trajectory (seconds)}} \\
    AndroidArena & 4.5 & 8.4 & 3.7 & 10.2 & 5.0 \\
    A3 (final state) & 3.7 & 9.5 & 4.6 & 13.4 & 6.9 \\
    A3 (essential) & 21.9 & 60.8 & 18.4 & 116.8 & 38.8 \\
    SPA-Bench & 11.9 & 18.8 & 8.0 & 47.0 & 15.0 \\
    AgentRewardBench & 5.6 & 17.6 & 5.3 & 21.9 & 11.6 \\
    Simple Baseline & 3.5 & 8.2 & 5.6 & 21.0 & 7.9 \\
    \bottomrule
    \end{tabular}
    \caption{Inference cost and latency of the 30 judges over the 931 trajectories.}
    \label{tab:cost}
    \end{table}

    \subsection{Self-judging}
    \label{app:self_judging}

    Two judge backbones share a model family with an agent that produced trajectories: Qwen2.5-VL-72B and GPT-5-mini. We test whether such a judge is more lenient toward its own family. Because agents differ in how detectable their failures are, a raw comparison of false-positive rates is confounded, so we use a difference-in-differences: the judge's false-positive rate on its own family's failed trajectories minus its rate on the other agents' failures, relative to the same gap averaged over the other four backbones. CIs come from the task-cluster bootstrap. Table~\ref{tab:self_judging} shows the result. Pooled over the six methods, the Qwen judge is harsher on its own family's failures and the GPT-5-mini judge shows no difference. Restricted to the simple baseline, the GPT-5-mini judge shows a borderline excess leniency toward the GPT-5-mini agent's failures ($+11.0$pp, estimated on 56 failed trajectories). Its predicted success rate for that agent exceeds the human rate by 4.6pp, which is within the 1.8--5.1pp range it shows for the other three agents, so the agent-level rates used in \S\ref{sec:meta_corr} are not visibly inflated.

    \begin{table}[h]
    \centering
    \footnotesize
    \setlength{\tabcolsep}{2pt}
    \begin{tabular}{lccc}
    \toprule
    \textbf{Judge backbone} & \textbf{Own} & \textbf{Other} & \textbf{Diff.-in-diff.} \\
    \midrule
    \multicolumn{4}{@{}l}{\emph{All six methods}} \\
    Qwen2.5-VL-72B & 13.4 & 31.7 & $-$8.8 {\scriptsize$[-11.6, -5.8]$} \\
    GPT-5-mini & 23.3 & 8.7 & +2.1 {\scriptsize$[-3.4, +7.8]$} \\
    \midrule
    \multicolumn{4}{@{}l}{\emph{Simple Baseline only}} \\
    Qwen2.5-VL-72B & 11.9 & 26.7 & $-$3.2 {\scriptsize$[-8.3, +2.2]$} \\
    GPT-5-mini & 37.5 & 10.7 & +11.0 {\scriptsize$[+0.8, +21.0]$} \\
    \bottomrule
    \end{tabular}
    \caption{Self-judging test. Own and Other: false-positive rate (\%) on failed trajectories of the judge's own model family and of the other agents. Diff.-in-diff.: the Own$-$Other gap relative to the same gap for the other four backbones (pp), with task-cluster bootstrap 95\% CIs.}
    \label{tab:self_judging}
    \end{table}

    \subsection{Operating points and thresholds}
    \label{app:thresholds}

    The evaluated methods emit binary verdicts. The verbalized confidence that the simple baseline elicits (high, medium, or low) is not a usable score: 95--100\% of verdicts are reported as high confidence on every backbone and none as low, in line with the overconfidence documented by \citet{xiong2024confidence}. We therefore study how a judge's operating point can be moved using two graded scores available in our outputs (Table~\ref{tab:thresholds}). First, A3 essential-states produces the fraction of essential states judged achieved. Used as a score, it has an AUROC of 0.82--0.88, and the method's default operating point, which requires all states, is not the most accurate one: with GPT-5-mini, a threshold of 0.8 raises accuracy from 78.1\% to 81.2\% (precision 85.6\%, recall 77.4\%, against 92.3\% and 63.8\% by default). Second, the number of the six methods that vote success on a given backbone has an AUROC of 0.92--0.96 and an AUPRC of 0.88--0.94. Raising the vote threshold trades recall for precision: with GPT-5-mini, three votes give precision 90.0\% and recall 93.3\%, a balanced point suited to evaluation, while five votes give 95.1\% and 67.4\%, a high-precision point closer to what reward modeling calls for.

    \begin{table*}[h]
    \centering
    \footnotesize
    \setlength{\tabcolsep}{5pt}
    \begin{tabular}{lcccccc}
    \toprule
     & \textbf{A3-ess.} & \multicolumn{5}{c}{\textbf{Method vote (of 6)}} \\
    \cmidrule(l){3-7}
    \textbf{Backbone} & \textbf{AUROC} & \textbf{AUROC} & \textbf{AUPRC} & \textbf{$\geq$3} & \textbf{$\geq$4} & \textbf{$\geq$5} \\
    \midrule
    Qwen & 0.82 & 0.92 & 0.88 & 79.9 / 97.0 & 83.7 / 93.7 & 87.1 / 87.8 \\
    GPT & 0.88 & 0.95 & 0.93 & 90.0 / 93.3 & 92.6 / 84.1 & 95.1 / 67.4 \\
    Gemini & 0.88 & 0.95 & 0.94 & 87.5 / 94.9 & 90.1 / 91.6 & 92.5 / 85.7 \\
    GLM & 0.86 & 0.94 & 0.92 & 82.0 / 96.1 & 85.7 / 92.2 & 90.4 / 81.1 \\
    Claude & 0.85 & 0.94 & 0.92 & 89.5 / 87.0 & 92.2 / 77.0 & 95.3 / 65.2 \\
    \bottomrule
    \end{tabular}
    \caption{Threshold analysis with two graded scores: the fraction of essential states achieved (A3 essential-states) and the number of methods voting success. The last three columns give precision / recall (\%) when a trajectory is accepted at that number of votes. Trajectories with a missing verdict from any method are excluded.}
    \label{tab:thresholds}
    \end{table*}

    \section{Training experiment details}
    \label{app:training}
    
    \paragraph{Base model and algorithm.}
    We use UI-TARS-7B-SFT as the base policy model and train with GRPO (Group Relative Policy Optimization) on 2$\times$H100 80GB GPUs, using the EasyR1 framework~\citep{zheng2025easyr1} built on verl~\citep{sheng2024hybridflow}. Training runs in 16 parallel Docker containers, each executing one AndroidWorld task instance at a time.
    
    \paragraph{Training data.}
    The training set is the easy task subset from AndroidWorld: 553 task instances across 38 parameterizable task templates and 14 apps. Each template is instantiated with different random seeds (2--16 seeds per template). The remaining 23 easy templates that produce identical instances regardless of seed are excluded from training but included in evaluation.
    
    \paragraph{Hyperparameters.}
    All conditions share identical hyperparameters (Table~\ref{tab:training_hparams}); only the reward source differs. Each training step processes 2 task instances with 8 rollouts each; the resulting 16 rollouts form one update batch. One episode iterates through all 553 instances (276 steps). Training proceeds for up to 300 steps with checkpoints every 50 steps.
    
    \begin{table}[h]
    \centering
    \small
    \begin{tabular}{ll}
    \toprule
    \textbf{Parameter} & \textbf{Value} \\
    \midrule
    Base model & UI-TARS-7B-SFT \\
    RL algorithm & GRPO \\
    Optimizer & AdamW (bf16) \\
    Learning rate & $1 \times 10^{-6}$ \\
    LR warmup & 5\% of steps (linear) \\
    KL coefficient & 0.05 \\
    Update epochs & 1 \\
    Clip ratio & [0.2, 0.3] (asymmetric) \\
    Max gradient norm & 1.0 \\
    Rollouts per task & 8 \\
    Tasks per step & 2 \\
    Max agent steps & 20 \\
    Sampling temperature & 1.0 (train), 0 (eval) \\
    \bottomrule
    \end{tabular}
    \caption{Training hyperparameters for on-policy RL experiments.}
    \label{tab:training_hparams}
    \end{table}

    \paragraph{Compute.}
    Each training run takes roughly two days on the 2$\times$H100 setup (${\approx}$100 GPU-hours per condition, ${\approx}$400 GPU-hours across the four conditions), excluding checkpoint evaluation. On the judge-evaluation side, the 6$\times$5 grid of Table~\ref{tab:main_results} comprises ${\approx}$28k trajectory-level judge evaluations, and our full logs across ablations and configuration sweeps total ${\approx}$57k evaluations, each involving one or more LLM calls. The final-frame and WebJudge runs (Appendices~\ref{app:final_frame} and~\ref{app:webjudge}) add ${\approx}$5.6k evaluations.

    \paragraph{Reward conditions.}
    The rule-based condition uses AndroidWorld's built-in Python verification scripts that programmatically check task completion by inspecting device state. The LLM judge conditions use our simple baseline judge (48 uniformly sampled screenshots, max long edge 600px, no UI metadata or agent reasoning) with three backbones: GPT-5-mini, GPT-5.2, and Qwen2.5-VL-72B. All produce binary rewards (1.0 for success, 0.0 for failure). The rule-based condition's reward coincides with the evaluation signal below, so its result is best read as an upper bound.
    
    \paragraph{Evaluation protocol.}
    All conditions are evaluated using the ground-truth rule-based checker, ensuring fair comparison. Evaluation covers all 116 AndroidWorld task templates across 3 seeds ($s \in \{7, 30, 1234\}$), with greedy decoding (temperature${}=0$). Success rate is computed as the fraction of successful tasks, averaged over 3 seeds (348 evaluations per checkpoint). \emph{Best checkpoint} refers to the checkpoint with the highest seed-averaged success rate for a condition.
    
    \paragraph{Full task-set results.}
    While Figure~\ref{fig:training_curves} reports the easy-set success rate, we also evaluate the same checkpoints on the complete 116-task suite. Taking the best success rate per condition (the same convention as for the easy set), the ordering is unchanged: rule-based 36.8\%, GPT-5-mini 30.2\%, GPT-5.2 27.9\%, Qwen 26.4\%. At the fixed final checkpoint (step 300), GPT-5.2 and Qwen tie at 25.9\%, so the precision comparison in \S\ref{sec:training} is specific to the best-checkpoint convention. Table~\ref{tab:full116} lists both conventions for every condition.

    \begin{table}[h]
    \centering
    \small
    \begin{tabular}{lcc}
    \toprule
    \textbf{Reward signal} & \textbf{Best} & \textbf{Final (step 300)} \\
    \midrule
    Rule-based       & 36.8 & 36.8 \\
    GPT-5-mini judge & 30.2 & 30.2 \\
    GPT-5.2 judge    & 27.9 & 25.9 \\
    Qwen judge       & 26.4 & 25.9 \\
    \bottomrule
    \end{tabular}
    \caption{Success rate (\%) on the complete 116-task AndroidWorld suite, averaged over 3 seeds: best checkpoint per condition vs.\ the final (step-300) checkpoint.}
    \label{tab:full116}
    \end{table}

\begin{figure*}[h]
    \centering
    \includegraphics[width=\textwidth]{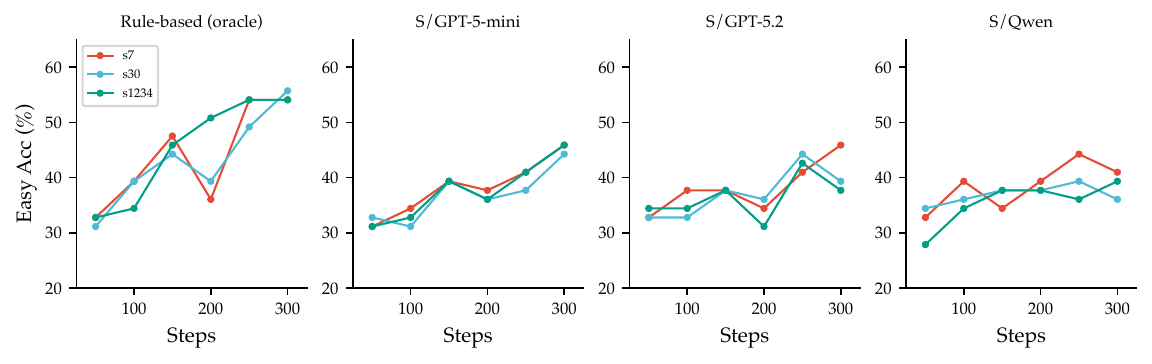}
    \caption{Per-seed easy-set accuracy for each reward condition. Seed variance is moderate; the ranking Rule-based $>$ GPT-5-mini $>$ GPT-5.2 $>$ Qwen is consistent across most seeds and steps.}
    \label{fig:training_per_seed}
    \end{figure*}

    \paragraph{Per-seed training curves.}
    Figure~\ref{fig:training_per_seed} shows the per-seed breakdown.
    
    \section{Failure analysis details}
    \label{app:failure}
    
    \subsection{Failure case examples}
    \label{app:failure_examples}
    
    Figures~\ref{fig:failure_examples_fn} and~\ref{fig:failure_examples_fp} show representative examples for each of the 7 failure categories, with a key trajectory screenshot and analysis.
    
    \newcommand{\failurecase}[6]{%
      \noindent
      \begin{minipage}[t]{0.15\textwidth}
        \vspace{0pt}
        \includegraphics[width=\textwidth,height=4cm,keepaspectratio]{#1}
      \end{minipage}%
      \hfill
      \begin{minipage}[t]{0.82\textwidth}
        \vspace{0pt}
        \textbf{#2} \\[1pt]
        \textbf{Task:} ``#3'' \quad \textbf{GT:} #4 \quad \textbf{Pred:} #5 \\[1pt]
        #6
      \end{minipage}
      \vspace{3pt}
      \hrule
      \vspace{3pt}
    }
    
    \begin{figure*}[h]
    \small
    
    \failurecase{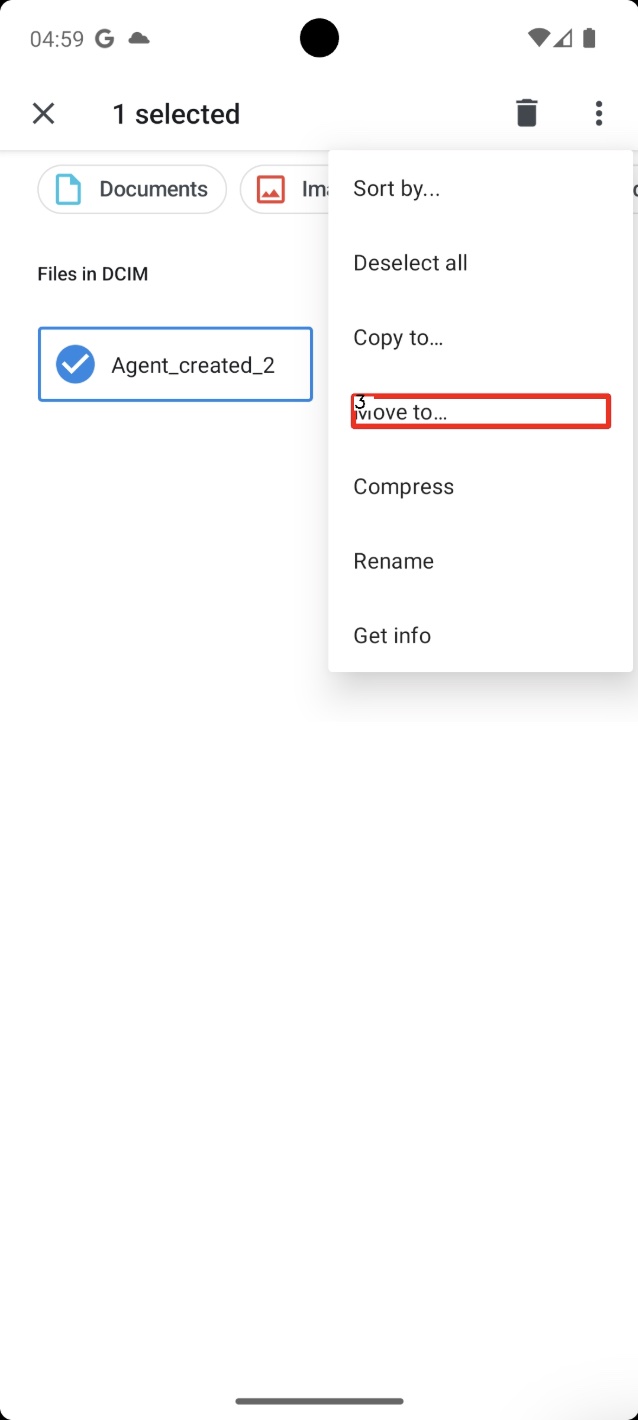}
    {FN: Last-frame anchoring (GPT: 12, Qwen: 1)}
    {Go to the DCIM folder in internal storage. Create a subfolder named Agent\_created.}
    {Success}{Failure}
    {The agent created the subfolder in an earlier step, but then navigated away. The final screenshot shows a different location, so judges anchored to the final frame miss the completion evidence.}
    
    \failurecase{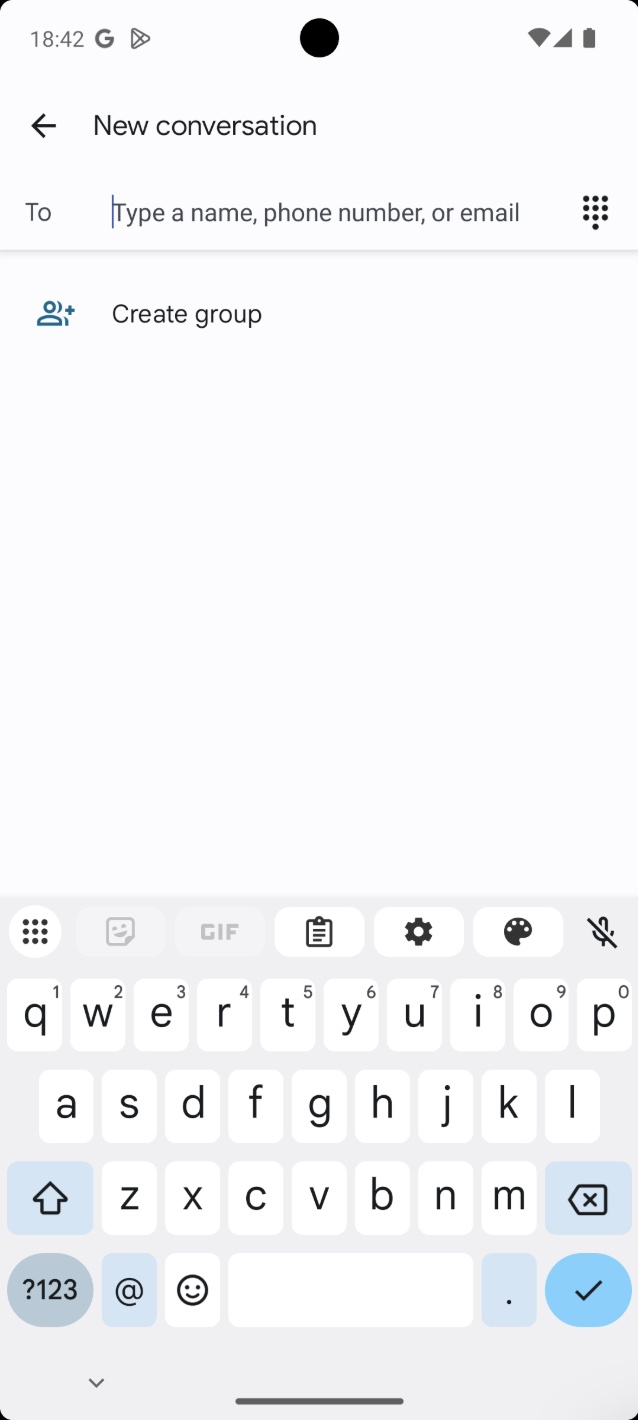}
    {FN: Unfamiliar success state (GPT: 9, Qwen: 4)}
    {Start chatting in message.}
    {Success}{Failure}
    {The agent reaches the ``New conversation'' screen with the recipient field active. Judges require a sent message, but the benchmark defines success as simply entering the chat composition flow.}
    
    \failurecase{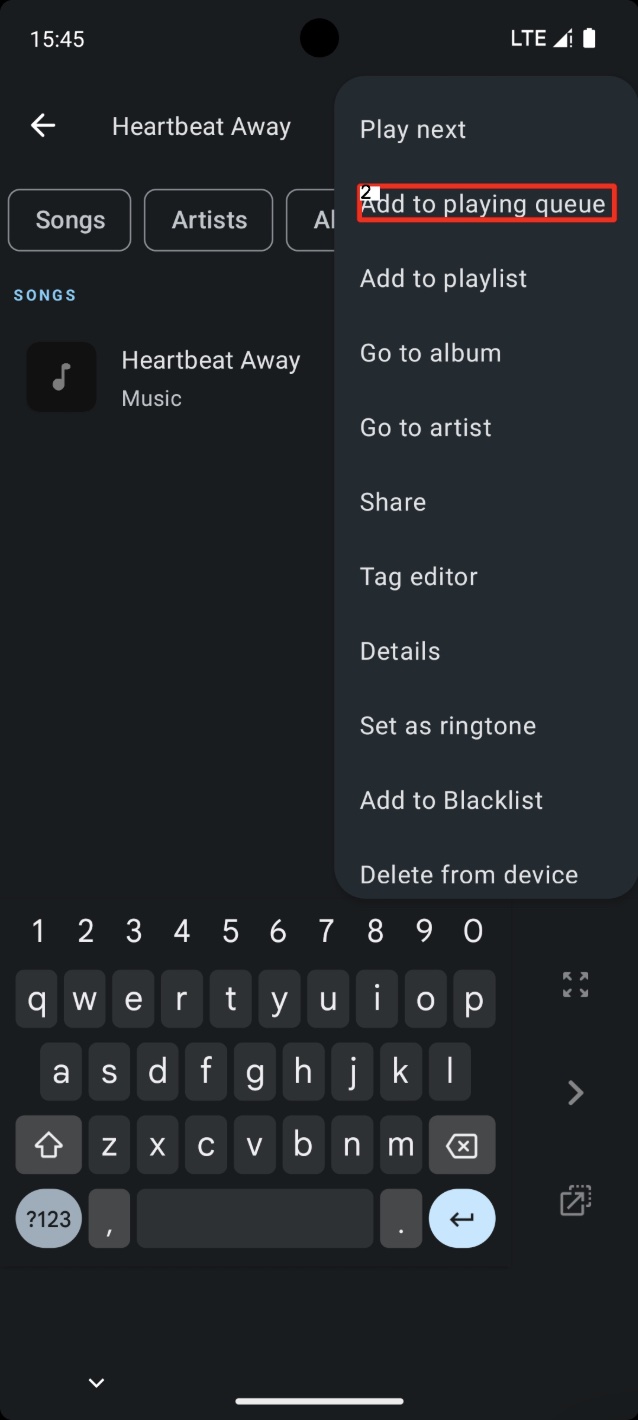}
    {FN: Invisible state change (GPT: 7, Qwen: 1)}
    {Add songs to the playing queue: Through the Storm, Hidden Paths, Forever Young, \ldots}
    {Success}{Failure}
    {The agent added songs via transient overflow-menu actions (``Add to queue''), but never opened the queue screen. No screenshot shows the resulting queue contents.}
    
    \failurecase{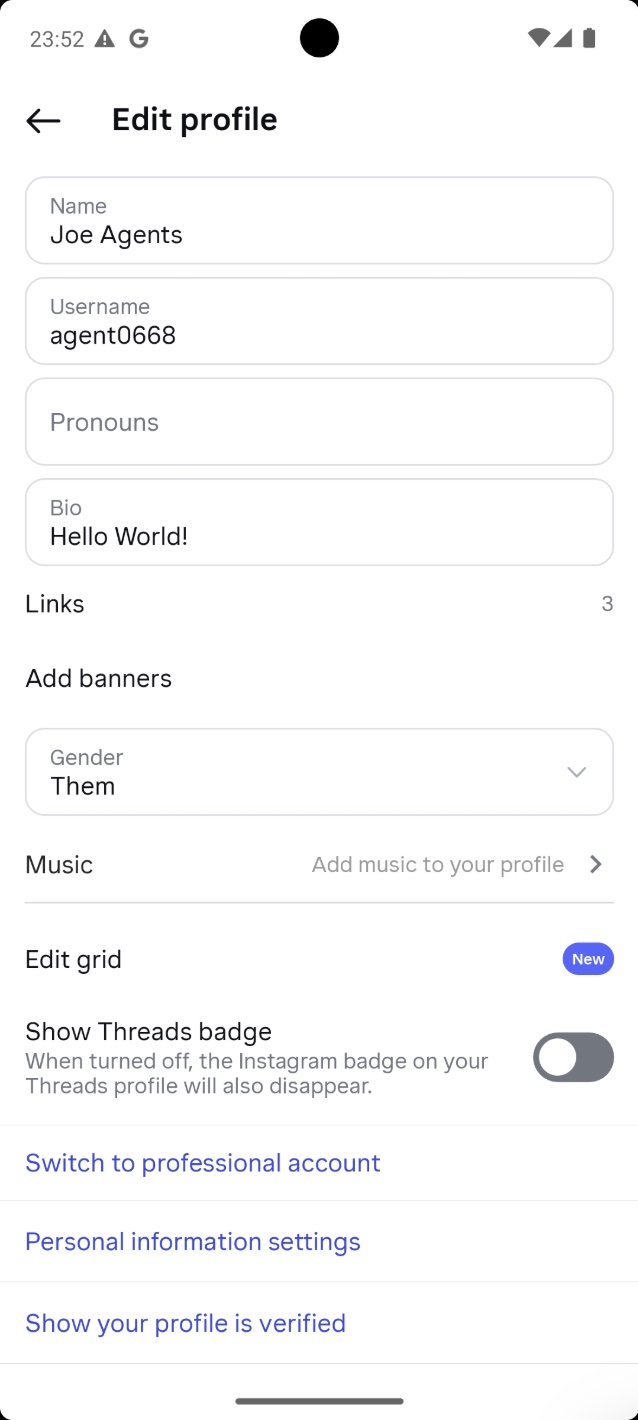}
    {FN: Verification gap (GPT: 2, Qwen: 1)}
    {Edit profile: add link, change gender to Custom, switch to private account.}
    {Success}{Failure}
    {Link and gender changes are evidenced by toasts in earlier steps, but the privacy toggle is never visually confirmed---no frame shows ``Private account'' being enabled.}
    
    \caption{Representative false negative failure examples (4 categories). Judges incorrectly predict failure despite task success.}
    \label{fig:failure_examples_fn}
    \end{figure*}
    
    \begin{figure*}[h]
    \small
    
    \failurecase{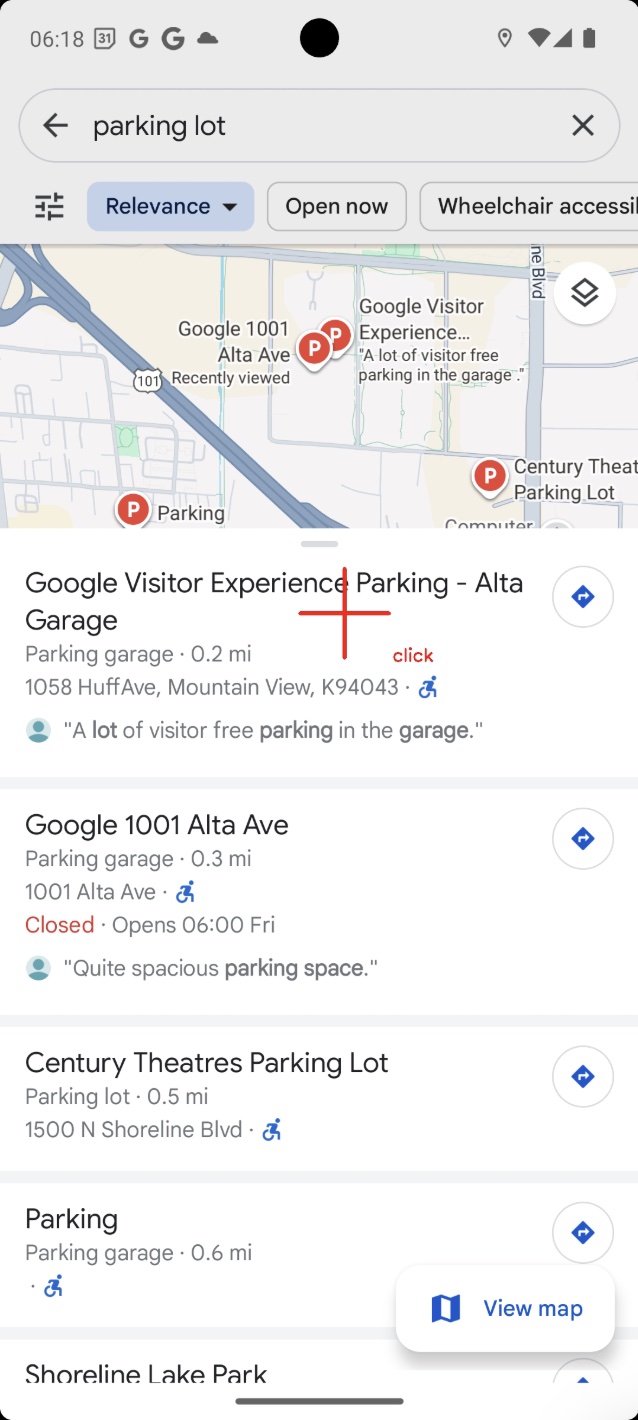}
    {FP: Constraint violation (GPT: 5, Qwen: 27)}
    {Find the nearest parking lot.}
    {Failure}{Success}
    {Judges accept opening any parking lot result from a Google Maps search. The task requires the \emph{nearest} one, but no judge verifies whether the selected result has the smallest distance.}
    
    \failurecase{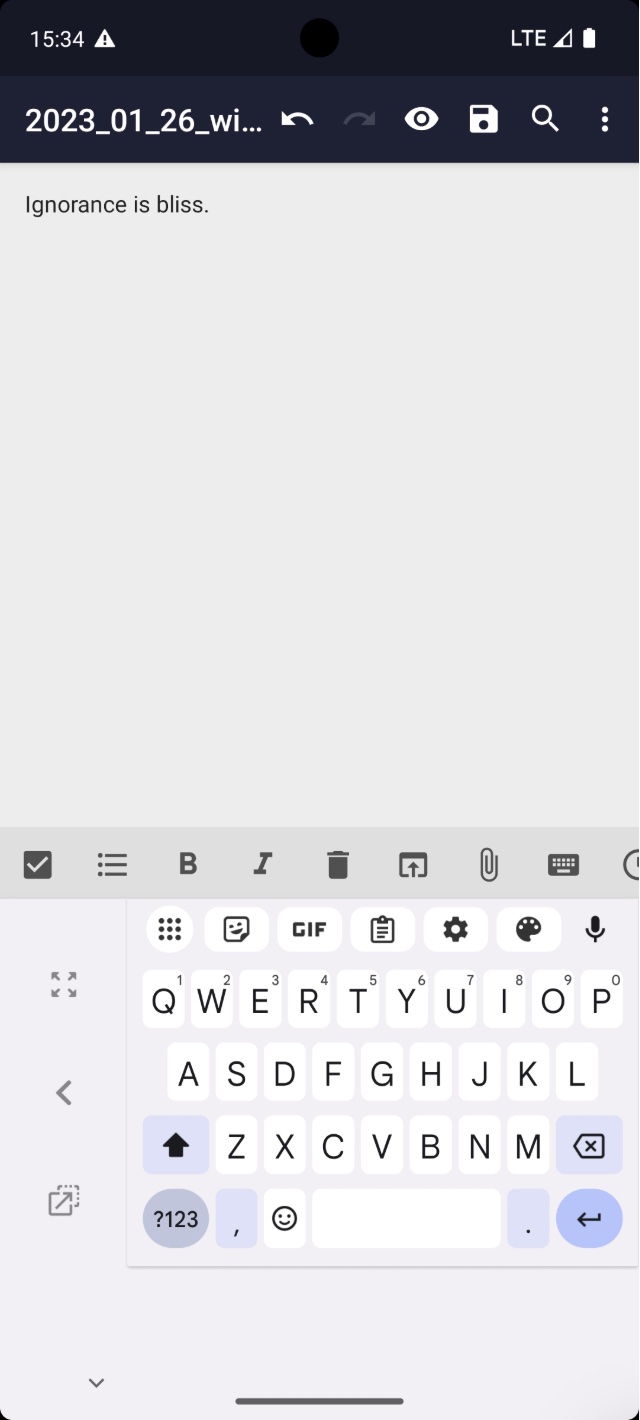}
    {FP: Partial completion (GPT: 5, Qwen: 27)}
    {Create a note in Markor named 2023\_01\_26\_wise\_yacht.md with text: Ignorance is bliss.}
    {Failure}{Success}
    {Judges see the filename and text in the editor and conclude success. However, the note was never saved---Markor shows an unsaved draft, and the file does not exist on the filesystem.}
    
    \failurecase{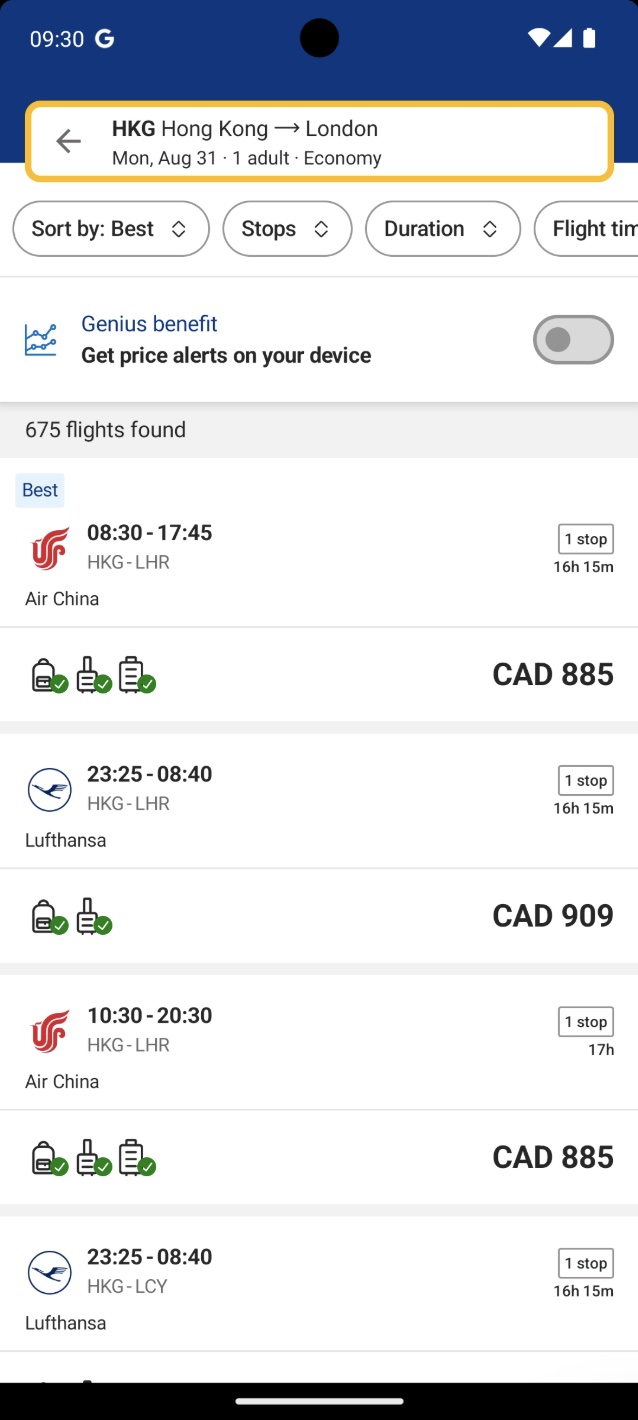}
    {FP: Surface UI match (GPT: 8, Qwen: 17)}
    {Search for one-way flights from Hong Kong to London on Aug 31. Which is the cheapest?}
    {Failure}{Success}
    {Judges see a flight results page with matching headers (HKG$\to$London, Aug 31, Economy) and assume success. The agent never identified the cheapest flight---the visual similarity was sufficient to fool all judges.}
    
    \caption{Representative false positive failure examples (3 categories). Judges incorrectly predict success despite task failure.}
    \label{fig:failure_examples_fp}
    \end{figure*}
    
    \subsection{Failure identification}
    
    For each LLM backbone (GPT-5-mini and Qwen2.5-VL), we run the 5 existing judge methods (excluding our simple baseline) on the 931 benchmark trajectories and identify cases where the judge prediction disagrees with human ground truth. We then compute the intersection of failures across methods: a trajectory is a \emph{hard-core failure} if at least 4 out of 5 methods produce the wrong prediction. This threshold ensures that the failures persist across methods and reflect cases that are fundamentally difficult for the given LLM backbone.
    
    \subsection{Root cause classification}
    
    For each hard-core failure case, we compile a structured dossier containing: (1)~the task instruction, (2)~the human ground-truth label and the error direction (FP or FN), (3)~the judge predictions and reasoning from all 5 methods, and (4)~key trajectory screenshots. We then classify each case into one of the following predefined error categories, based on examining the trajectory evidence and judge reasoning:
    
    \paragraph{False negative categories (judges miss true success):}
    \begin{itemize}
        \item \textbf{Last-frame anchoring}: The judge over-relies on the final screenshot and misses evidence of task completion from earlier steps.
        \item \textbf{Invisible state change}: The task was completed but the change is not visually evident (\eg, a setting toggled internally, an item deleted from a database).
        \item \textbf{Verification gap}: The trajectory lacks explicit visual confirmation for irreversible actions (\eg, delete, post, toggle) despite a consistent final state.
        \item \textbf{Unfamiliar success state}: The judge applies overly strict criteria misaligned with the benchmark's definition of success.
    \end{itemize}
    
    \paragraph{False positive categories (judges accept true failures):}
    \begin{itemize}
        \item \textbf{Surface UI match}: The judge concludes success based on superficial visual similarity (\eg, a screen that looks like the target but is not).
        \item \textbf{Partial completion}: Only a subset of the task requirements is met; the judge overlooks the remaining components.
        \item \textbf{Constraint violation}: The judge misses task-specific constraints (\eg, ``nearest,'' ``cheapest,'' a specific date or quantity).
    \end{itemize}
    
    Using these categories, we classify 48 hard-core failures for the GPT-5-mini set and 78 for the Qwen set. The full distributions are shown in Figure~\ref{fig:failure_taxonomy}.

\subsection{Taxonomy reliability}
\label{app:taxonomy_reliability}

Two raters independently categorized a random sample of 30 hard-core failure cases (10 false negatives, 20 false positives) into the seven root-cause categories, given the task instruction, the ground-truth outcome, the trajectory, the judges' reasoning, and the written category definitions. The raters agreed on 29 of 30 cases (96.7\%; Cohen's $\kappa = 0.957$; false-negative subset 10/10, $\kappa = 1.000$; false-positive subset 19/20, $\kappa = 0.924$). The single disagreement is a borderline case for which two categories plausibly apply (surface UI match vs.\ constraint violation). We attribute the high agreement to the categories being defined in terms of concrete, observable behaviors.

\section{Use of AI Assistants}
\label{app:ai_use}
 
We used AI Assistants for writing assistance: editing prose for clarity and consistency of register, and suggesting rewrites of individual passages.
The authors reviewed every suggestion and applied them selectively.
We also used AI Assistants for coding and analysis support, and the authors reviewed and verified the resulting code and results.
Literature search was performed by the authors without AI assistance.
All research ideas, experimental design, analyses, results, and claims are the authors' own; the authors verified all AI-suggested text and take full responsibility for the content of this paper. 

\end{document}